\documentclass[preprint,review,12pt]{elsarticle}

\usepackage{amssymb}
\usepackage{graphicx}
\usepackage{url}
\usepackage{multirow}
\usepackage{booktabs}
\usepackage{subfigure}
\usepackage{picins}

\journal{Neurocomputing}

\begin{document}

\begin{frontmatter}



\title{An Effective Unconstrained Correlation Filter\\
and Its Kernelization for Face Recognition}


\author{Yan Yan$^{1}$}
\author{Hanzi Wang$^{1}$\corref{cor1}}
\cortext[cor1]{Corresponding author.}
\author{Cuihua Li$^{1}$}
\author{Chenhui Yang$^{1}$}
\author{Bineng Zhong$^{1,2}$}

\address{$^{1}$School of Information Science and Technology, Xiamen University, China\\
$^{2}$Department of Computer Science and Engineering, Huaqiao University, China}

\begin{abstract}
In this paper, an effective unconstrained correlation filter called Unconstrained Optimal Origin Tradeoff Filter (UOOTF) is presented and applied to robust face recognition. Compared with the conventional correlation filters in Class-dependence Feature Analysis (CFA), UOOTF improves the overall performance for unseen patterns by removing the hard constraints on the origin correlation outputs during the filter design.
To handle non-linearly separable distributions between different classes, we further develop a non-linear extension of UOOTF based on the kernel technique.
The kernel extension of UOOTF allows for higher flexibility of the decision boundary due to a wider range of non-linearity properties.
Experimental results demonstrate the effectiveness of the proposed unconstrained correlation filter and its kernelization in the task of face recognition.
\end{abstract}

\begin{keyword}
Unconstrained correlation filter\sep Kernel method\sep Class-dependence feature analysis (CFA)\sep Face recognition


\end{keyword}

\end{frontmatter}


\newpage
\section{Introduction}

Over the past few decades, increasing interest in biometrics has led
to rapid improvements in biometric
technologies \cite{Jain2004}.
Various biometric technologies
are available for identifying or verifying an individual.
Face recognition, in particular, has attracted much attention due to its non-intrusive nature and important role in the areas of access control, security, video surveillance, and so on \cite{Zhao2003}. However, face recognition is a very challenging task in practice due to great variations in facial appearance caused by pose, illumination, expression, etc. Particularly, in real-world applications, face recognition often encounters the small sample size (SSS) problem \cite{Jain1997,Tan2008}, where the training samples of subjects are very limited while the dimensionality of face data is high.

A variety of face recognition algorithms have been developed so far \cite{Zhao2003}. Among them, the appearance-based methods are one of the well-studied techniques where a face is usually represented as a high-dimensional vector. To overcome the problems incurred by high dimensionality,
subspace learning methods \cite{Li2010}, which aim to find linear/non-linear mappings, are used to project the
high-dimensional data onto the low-dimensional subspace.
Typical subspace learning methods include Principal Component Analysis (PCA) \cite{Turk1991}, Linear Discriminant Analysis (LDA) \cite{Belhumeur1997,Yang2005} and related methods \cite{Tao2007a, Tao2007b, Tao2009}, Locality Preserving Projections (LPP) \cite{He2005}, Non-negative Matrix Factorization (NMF) \cite{Lee1999}, and Class-dependence Feature Analysis (CFA) \cite{Xie2005a,Kumar2006}.

The projection axis obtained by the traditional subspace learning methods, such as PCA, LDA and LPP, is used to preserve the dominant data information or discriminate all the classes. One common problem of these methods is that they are not able to effectively discriminate classes close to each other since large class distances are often overemphasized during training. The resulting transformed subspace can preserve the distances of well-separated classes, while causing overlaps between neighboring classes. Tao et al. \cite{Tao2007b,Tao2009} proposed a new criterion based on the maximization of the geometric mean of the divergences (MGMD) between different pairs of classes for subspace selection, which reduces the class separation problem. Recently, NMF \cite{Lee1999} and its variants \cite{Guan2011,Guan2012a,Guan2012b} were developed as new subspace learning methods. Based on the fact that many real-world data, such as images or videos, are non-negative, NMF-related methods force the non-negativity constraints in factorization. This non-negativity constraints are consistent with the psychological evidence of parts-based representations for human perception. However, the computational complexity of NMF-related methods is high for large training data.

        \begin{figure*}[tbh!]
         \centering
            \includegraphics[width=8cm,height=6.5cm]{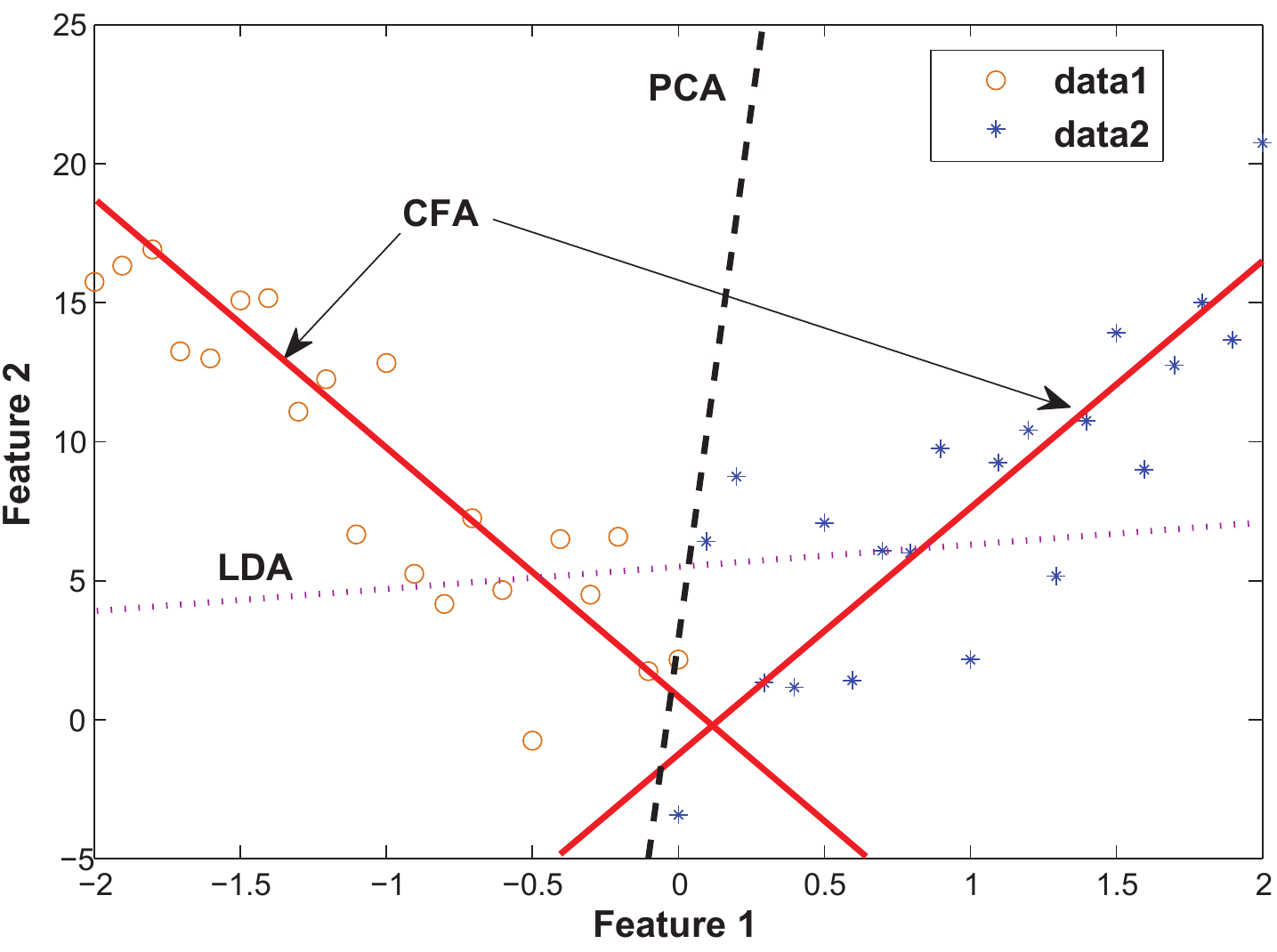}
         \caption{A comparison between different subspace learning methods for a two-class problem. The first projection axes for PCA and LDA are shown in black and purple, respectively. Two projection axes for CFA are given in red. Note that CFA obtains a projection axis for each class while the projection axes of PCA and LDA try to preserve or discriminate all the classes.
                 }
         \label{FIG:Example}%
        \end{figure*}

Compared with the above methods, the projection axis obtained by CFA is used to distinguish one specific class from the other classes (see Fig. \ref{FIG:Example} for an illustration). Besides, the traditional methods \cite{Turk1991,Belhumeur1997,Yang2005,Tao2007a,He2005,Lee1999} often employ features derived from the space domain, while CFA uses features derived from the frequency domain \cite{Xie2005a,Kumar2006}. The key step in CFA is the design of the correlation filters. Phase information which contains the structural information for human perception is directly modeled by the correlation filters in CFA \cite{Kumar2006}. What is more, the correlation filters offer some desirable properties, such as graceful degradation, shift-invariance, and closed-form solutions  \cite{Xie2005a,Kumar2006,Xie2005b,Yan2008a,Yan2008b,Han2010,Bolme2011}.

The original CFA \cite{Xie2005a,Kumar2006} designs the correlation filters by using the 2D Fourier transforms of images. For simplicity, we call the original CFA as 2D-CFA. In our previous work \cite{Yan2008a}, a tensor correlation filter based CFA method (TCF-CFA), which generalizes the original CFA by encoding the image data as tensors, was proposed. It has been shown that TCF-CFA can be derived in a similar way as 2D-CFA, which is a special case of TCF-CFA when the image data are encoded as 2nd-order tensors (i.e., image matrices). Moreover, the commonly-used correlation filters in TCF-CFA have the same form as those in 2D-CFA. In this study, we mainly focus on the 1D correlation filter based CFA (1D-CFA), since the previous research has demonstrated the great success by considering the image data as vectors \cite{Turk1991,Belhumeur1997,Yang2005,He2005} and experimental results have already shown that 1D-CFA and 2D-CFA can achieve similar performance \cite{Yan2008a,Yan2008b}.

There have been some widely-used correlation filters existing in the literature. For example, Mahalanobis et al. \cite{Mahalanobis1987} proposed the Minimum Average Correlation Energy (MACE) filter. The objective function of the MACE filter is to minimize the average energy of the correlation plane outputs while satisfying the correlation peak amplitude constraints. The MACE filter emphasizes high spatial frequencies to produce sharp correlation peaks, which makes it very sensitive to noise. Kumar \cite{Kumar1986} derived the Minimum Variance Synthetic Discriminant Function (MVSDF) filter, which minimizes the correlation output noise variance while satisfying the correlation peak amplitude constraints. The MVSDF filter focuses on low spatial frequencies to reduce noise. OTF (Optimal Tradeoff Filter) \cite{Refregier1990} combines the MACE filter and the MVSDF filter to produce sharp correlation peaks and suppress noise. OEOTF (Optimal Extra-class  Output Tradeoff Filter) \cite{Yan2008b} was proposed to optimize the extra-class correlation outputs at the origin of the correlation plane. Besides, the unconstrained correlation filters, such as the Unconstrained OTF filter (UOTF) \cite{Kumar1999,Mahalanobis1994}, are designed to maximize the average correlation height instead of enforcing the hard constraints on the outputs of correlation filters.

The traditional correlation filters, such as MACE \cite{Mahalanobis1987}, MVSDF \cite{Kumar1986}, OTF \cite{Refregier1990}, and OEOTF\cite{Yan2008b}, assume that the distortion tolerance of a filter could be controlled by explicitly specifying desired correlation peak values for the training images. As a matter of fact, the overall performance becomes worse if one enforces the hard constraints on the correlation peak values during training. Relaxing the hard constraints by using the unconstrained form could improve the overall performance for unseen patterns \cite{Mahalanobis1994}. Unfortunately, experimental results on face recognition show that the direct use of UOTF \cite{Kumar1999,Mahalanobis1994} is not desirable for feature extraction. The reason is that the design criterion of UOTF is not optimized for feature extraction in CFA. Thus, it motivates us to design an effective unconstrained correlation filter which is in consistence with the feature extraction process of CFA.

In this paper, we propose a novel and effective unconstrained correlation filter, called Unconstrained Optimal Origin Tradeoff Filter (UOOTF), to extract the effective discriminative features in CFA. Furthermore, to handle non-linearly separable distributions between different face classes, we also develop a nonlinear extension of UOOTF (called KUOOTF) based on the kernel technique \cite{Mika1999,Liu2002}. As far as we know, very few work concerns the design of unconstrained correlation filters in the CFA framework.

In summary, the main contribution in this paper is a novel unconstrained correlation filter (i.e., UOOTF) for effective feature extraction. UOOTF has three main advantages: 1) UOOTF overcomes the overfitting problem of the traditional UOTF by emphasizing the origin correlation outputs;
2) UOOTF provides a better generalization capability for unseen patterns by removing the hard constraints during the filter design; 3) UOOTF can be easily extended to the kernel form (i.e., KUOOTF) to deal with the non-linear
structure of the class distribution.

The rest of this paper is organized as follows. In Section 2, the
details of the proposed unconstrained correlation filter (UOOTF) are presented.
In Section 3, we show how to extend UOOTF to its kernel form by using the kernel technique.
In Section 4, the proposed methods in the task of face recognition on several popular face databases are evaluated. Finally, we provide some
concluding remarks in Section 5.

\section{Unconstrained Optimal Origin Tradeoff Filter}
In this section, we begin with briefly introducing 1D-CFA \cite{Yan2008a,Yan2008b} in Section 2.1, since our method mainly focuses on the correlation filter design in the 1D-CFA framework. Then, the proposed unconstrained correlation filter design is presented in Section 2.2. In Section 2.3, we discuss the distinctions between different correlation filters.
\subsection{1D-CFA}
\begin{figure*}[tbh!]
         \centering
            \includegraphics{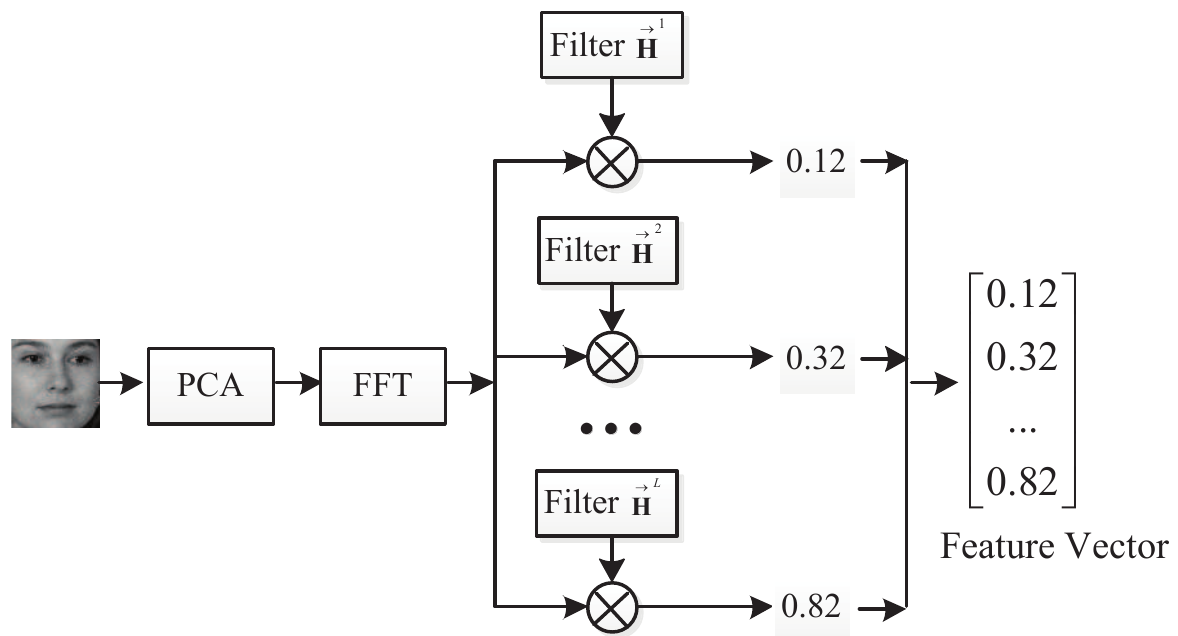}
         \caption{Feature extraction in 1D-CFA. Note that FFT is the Fast Fourier Transform which effectively computes the discrete Fourier transform.
                 }
         \label{FIG:Framework}%
        \end{figure*}

Compared with 2D-CFA where the correlation filters are designed in the two-dimensional image space, 1D-CFA not only achieves similar accuracy, but also has much lower computational complexity \cite{Yan2008b,Han2010}. During the training stage, the 1D-CFA projection vectors (i.e., the correlation filters) are generated and used for feature extraction. More specifically, face images are firstly represented as high-dimensional data (e.g. the pixel intensities \cite{Turk1991} or Gabor features \cite{Liu2004}). Then, PCA is used to perform dimensionality reduction. In the PCA subspace, the correlation filters are obtained by using the 1D Fourier transforms of the low-dimensional features. Finally, a bank of class-dependence correlation filters is trained for feature extraction, as shown in Fig. \ref{FIG:Framework}. Note that only the origin correlation outputs are used to form the feature in 1D-CFA.

\subsection{Unconstrained Correlation Filter Design}

 The traditional correlation filters \cite{Yan2008b,Mahalanobis1987,Kumar1986,Refregier1990} in 1D-CFA are based on the assumption that the correlation peak amplitude should satisfy a specified value (i.e., the origin correlation outputs are restricted to $1$ for a specific class and 0 for the others). However, the overall performance of those filters can become worse for unseen patterns if the correlation peak values are constrained to some specified constant values during the filter design, which motivates us to design the filter in the unconstrained form.

 UOTF is a traditional unconstrained correlation filter. The design criterion of UOTF is to: (i) minimize the average energy and noise variance of the whole correlation plane for all the samples; and (ii) maximize the origin correlation outputs for the intra-class samples. However, the minimization of (i) cannot guarantee that the origin correlation outputs for the extra-class samples (used to form the feature) are minimal. As a result, although UOTF \cite{Mahalanobis1994} tries to overcome the generalization problem by removing the hard constraints of OTF, UOTF fails in 1D-CFA (see Section 4 for the experimental results). Therefore, in this paper, we propose to directly optimize the origin correlation outputs and take the extra-class samples and intra-class samples into respective considerations during the filter design.

In the following, we describe the details of the proposed UOOTF.
For the clarity of presentation, vectors are denoted by an arrow on top of the alphabet. Upper case symbols refer to quantities in the frequency plane terms while lower case symbols represent quantities in the space domain.

1D-CFA designs a correlation filter for each class. Let the filter trained for the $l$-th class be $\vec{h}^{l}$, and $\vec{o}^{~l}_{i}$ be the output of  $\vec{h}^{l}$ in response to $\vec{y}_{i}$. We have
\begin{equation}
\label{EQ:1}
\vec{o}^{~l}_{i}(n) = {\vec{y}_{i}}(n) \odot {\vec{h}}^{l}(n),
\end{equation}
where  $\odot$ is a correlation function;  $\vec{y}_{i}$ is the low-dimensional PCA feature for the $i$-th training image; $n$ is the feature index in the spatial domain.

Equation (\ref{EQ:1}) can be expressed in the frequency domain by using the 1D Fourier transform as follows:
\begin{equation}
\label{EQ:2}
\vec{o}^{~l}_{i}(n) = \sum_{k=0}^{p-1}\vec{\bf{Y}}_{i}(k)^{*}\cdot \vec{\bf{H}}^{l}(k)e^{\frac{j2\pi kn}{p}}. 
\end{equation}
Here, $\vec{\bf{Y}}_{i}$  and   $\vec{\bf{H}}^{l}$ represent the 1D Fourier transforms of  $\vec{y}_{i}$ and $\vec{h}^{l}$, respectively; $p$ is the reduced dimensionality of the PCA subspace; $k$ is the feature index in the frequency domain; `$*$' denotes the conjugate operator.
According to (\ref{EQ:2}), the origin correlation output ($n=0$) is the inner product of the input signal and the correlation filter in the frequency domain.

        \begin{figure*}[tbh!]
         \centering
            \includegraphics{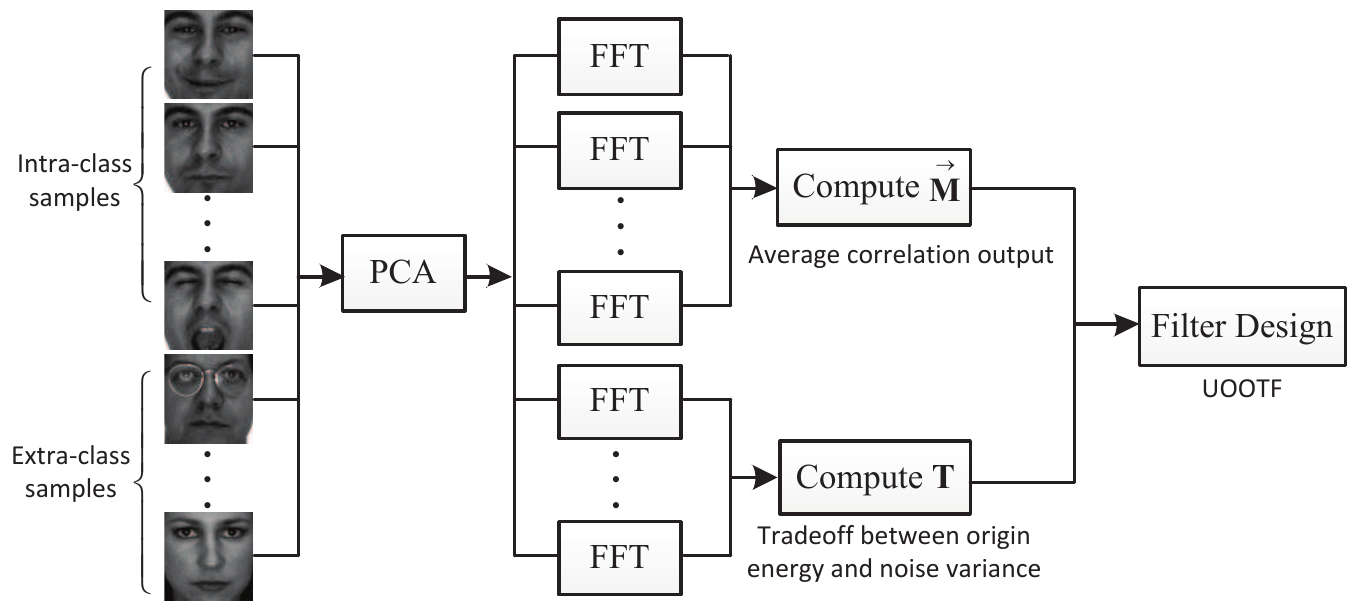}
         \caption{Framework of the UOOTF design.
                 }
         \label{FIG:FilterDesign}%
        \end{figure*}
The framework of the UOOTF design is shown in Fig. \ref{FIG:FilterDesign}. For the extra-class samples, UOOTF tries to balance the tradeoff between the origin correlation output energy and the origin correlation output noise variance. It can be derived by minimizing the weighted sum of the origin energy $|\vec{o}^{~l}_{i}(0)|^{2}$ and the origin noise variance $|\vec{n}^{~l}_{i}(0)|^{2}$ for the extra-class samples, which is expressed as
\begin{eqnarray}
\label{EQ:3}%
& &\min_{\vec{\bf{H}}^{l}}\omega_{s}\left( \frac{1}{N_{el}}\sum_{i=1}^{N_{el}}|\vec{o}^{~l}_{i}(0)|^{2}\right) + \omega_{n}\left(\frac{1}{N_{el}}\sum_{i=1}^{N_{el}}|\vec{n}^{~l}_{i}(0)|^{2}\right) \nonumber \\
&=& \min_{\vec{\bf{H}}^{l}}\omega_{s}\left( \frac{1}{N_{el}}\sum_{i=1}^{N_{el}}|\vec{\bf{Y}}_{i}^{\mathrm{E}+}\vec{\bf{H}}^{l}|^{2}\right) + \omega_{n}\left(\frac{1}{N_{el}}\sum_{i=1}^{N_{el}}|\vec{\bf{N}}_{i}^{\mathrm{E}+}\vec{\bf{H}}^{l}|^{2}\right)
\nonumber \\
&=& \min_{\vec{\bf{H}}^{l}}\omega_{s} {\vec{\bf{H}}^{l+}}{\mathbf{R}}_{\mathrm{Y}}^{l}\vec{\bf{H}}^{l} + \omega_{n}\vec{\bf{H}}^{l+}{\mathbf{C}}{\vec{\mathbf{H}}}^{l},
\end{eqnarray}
where ${\mathbf{R}}_{\mathrm{Y}}^{l} = 1/N_{el}\sum_{i=1}^{N_{el}}\vec{\bf{Y}}_{i}^{\mathrm{E}}\vec{\bf{Y}}_{i}^{\mathrm{E}+}$, and $\vec{\bf{Y}}_{i}^{\mathrm{E}}$ $( i = 1,\cdots,N_{el})$ is the 1D Fourier transform of the extra-class sample for the $l$-th class. $\mathbf{C}=$ $1/N_{el}\sum_{i=1}^{N_{el}}\vec{\bf{N}}_{i}^{\mathrm{E}}\vec{\bf{N}}_{i}^{\mathrm{E}+}$,
 and $\vec{\bf{N}}_{i}^{\mathrm{E}}$ $( i = 1,\cdots,N_{el})$ is the 1D Fourier transform of the extra-class noise sample for the $l$-th class; $\mathbf{C}$
 is usually set as a diagonal matrix whose diagonal elements represent the noise power spectral density (In fact, $\mathbf{C}$ can also be viewed as a regularization term); `$+$' represents the conjugate transpose; $\omega_{s}$ and $\omega_{n}$ $(0\leq\omega_{s},\omega_{n}\leq1)$  are the tradeoff parameters; $N$ is the number of all the training samples and $N_{l}$ is the number of training samples for the $l$-th class; $N_{el} = N-N_{l}$ denotes the number of extra-class training samples for the $l$-th class.

For the intra-class samples, we try to maximize the average origin correlation output, which is given by
\begin{equation}
\label{EQ:4}
\max_{\vec{\bf{H}}^{l}}(\frac{1}{N_{l}}\sum_{i=1}^{N_{l}}\vec{\bf{Y}}_{i}^{\mathrm{I}+}\vec{\bf{H}}^{l}) = \max_{\vec{\bf{H}}^{l}}(\vec{\mathbf{M}}^{l+}\vec{\mathbf{H}}^{l}),
\end{equation}
where $\vec{\bf{M}}^{l}=1/N_{l}\sum_{i=1}^{N_{l}}\vec{\bf{Y}}_{i}^{\mathrm{I}}$ is the average of all the intra-class samples for the $l$-th class, and $\vec{\bf{Y}}_{i}^{\mathrm{I}}$ $( i = 1,\cdots,N_{l})$ is the 1D Fourier transform of the intra-class sample for the $l$-th class.

By combining (\ref{EQ:3}) and (\ref{EQ:4}),  we have the following optimization criterion:
\begin{eqnarray}
\label{EQ:5}
J(\vec{\bf{H}}^{l})   & = & \frac{|\vec{\mathbf{M}}^{l+}\vec{\mathbf{H}}^{l}|^{2}}{\omega_{s}{\vec{\bf{H}}^{l+}}{\mathbf{R}}_{\mathrm{Y}}^{l}\vec{\bf{H}}^{l} + \omega_{n}\vec{\bf{H}}^{l+}{\mathbf{C}}{\vec{\mathbf{H}}}^{l}}  \nonumber \\
& =&  \frac{\vec{\mathbf{H}}^{l+}\vec{\mathbf{M}}^{l}\vec{\mathbf{M}}^{l+}\vec{\mathbf{H}}^{l}}{{\vec{\bf{H}}^{l+}}(\omega_{s} {\mathbf{R}}_{\mathrm{Y}}^{l} + \omega_{n}{\mathbf{C}}){\vec{\mathbf{H}}}^{l}}.
\end{eqnarray}

The unconstrained correlation filter UOOTF can then be derived by maximizing the criterion function $J(\vec{\bf{H}}^{l}) $, i.e.,
\begin{equation}
\label{EQ:6}
\vec{\mathbf{H}}^{l} = \rm{arg} \max_{{\vec{\mathbf{H}}}^{\it{l}}}{J({\vec{\mathbf{H}}}^{\it{l}})}.
\end{equation}

By using the Lagrange multiplier method \cite{Boyd2004}, it is easy to derive the following closed-form solution of UOOTF:
\begin{equation}
\label{EQ:7}
\vec{\mathbf{H}}^{l} = (\mathbf{T}^{l})^{-1}\vec{\mathbf{M}}^{l},
\end{equation}
where $\mathbf{T}^{l} = \omega_{s}{\mathbf{R}}_{\mathrm{Y}}^{l} + \omega_{n}{\mathbf{C}}$.

Since the hard constraints are removed during the filter design, the peak values at the origin vary for classes. To overcome the scale differences for different correlation filters, we normalize the feature by using a simple strategy as follows:
\begin{equation}
\label{EQ:8}
\vec{\rm{x}}_{\rm{n}} = \frac{\vec{\rm{x}}}{\max(\vec{\rm{x}})}.
\end{equation}
Here,  $\max(\vec{\rm{x}})$ returns the maximum value in the vector $\vec{\rm{x}}$; $\vec{\rm{x}}_{\rm{n}}$ is the normalized feature.

In Algorithm 1, we give the outline of the proposed UOOTF based 1D-CFA for face recognition.
\begin{table}
\centering
\vspace{3mm}
\noindent
\scalebox{0.85}{
\begin{tabular}
{p{348pt}}
\hline
\textbf{Algorithm 1} UOOTF based 1D-CFA for face recognition \\
\hline
\vspace{0.05mm}
\textbf{Input:} Query image $\vec{p}_{\rm{q}} \in \Re^{m\times1}$, and training data matrix $\rm{D} \in \Re^{\it{m}\times}$$^N$   with $L$ classes, where $m$ is the
        dimensionality of the face feature.\\
\textbf{Output:} The class label of the query image.
\vspace{2mm}
\\
\hline
\vspace{0.1mm}
\textbf{Training Stage:}\\
\emph{Step 1}: Project the training data matrix  $\rm{D} \in$ $\Re^{m\times}$$^N$  onto the PCA subspace to obtain the low-dimensional feature matrix $\rm{Y} \in$$ \Re^{p\times}$$^N$  and the corresponding 1D Fourier transform matrix $\mathbf{Y} \in \Re^{p\times }$$^N$ .\\
\emph{Step 2}: Do for $l$ = 1,$\cdots$, $L$: \\
 \quad \quad \quad 2.1 Calculate the tradeoff $\mathbf{T}^{l}$ using the extra-class samples of the $l$-th class;\\
 \quad \quad \quad 2.2 Calculate the average value $\vec{\mathbf{M}}^{l}$  using the intra-class samples of the $l$-th class;\\
 \quad \quad \quad 2.3 Design the correlation filter $\vec{\bf{H}}^{l}$ by (\ref{EQ:7}).\\
\emph{Step 3}: Construct the projection matrix $\mathbf{P}=[\vec{\bf{H}}^{1},\cdots,\vec{\bf{H}}^{L}]$ .\\
\emph{Step 4}: Compute the feature matrix $\rm{X} = \mathbf{P}^{T}\mathbf{Y}$.\\
\emph{Step 5}: Normalize each column of the feature matrix X based on (\ref{EQ:8}) to obtain the normalized feature matrix $\rm{X}_{\rm{n}}$ .\\
\vspace{0.01mm}
\textbf{Testing Stage:}\\
\emph{Step 1}:  Project the query face $\vec{p}_{\rm{q}}$ onto the PCA subspace to obtain the low-dimensional feature $\vec{y}_{\rm{q}} \in  \Re^{p\times1}$ and the corresponding 1D Fourier transform $\vec{\mathbf{Y}}_{\rm{q}} \in \Re^{p \times 1}$  .\\
\emph{Step 2}: Compute the feature $\vec{\rm{x}}_{\rm{q}} = \mathbf{P}^{\rm{T}}\vec{\mathbf{Y}}_{\rm{q}}$.\\
\emph{Step 3}: Normalize the feature  $\vec{\rm{x}}_{\rm{q}}$ based on (\ref{EQ:8}) to obtain $\vec{\rm{x}}_{\rm{qn}}$ .\\
\emph{Step 4}: Assign the class label to the query image $\vec{p}_{\rm{q}}$  by using the nearest neighbor classifier based on $\vec{\rm{x}}_{\rm{qn}}$  and  $\rm{X}$$_{\rm{n}}$.
\vspace{2mm}\\
\hline
\end{tabular}
}
\end{table}

\vspace{2mm}
\subsection{Discussions}
\begin{figure*}[tbh!]
         \centering
            \includegraphics{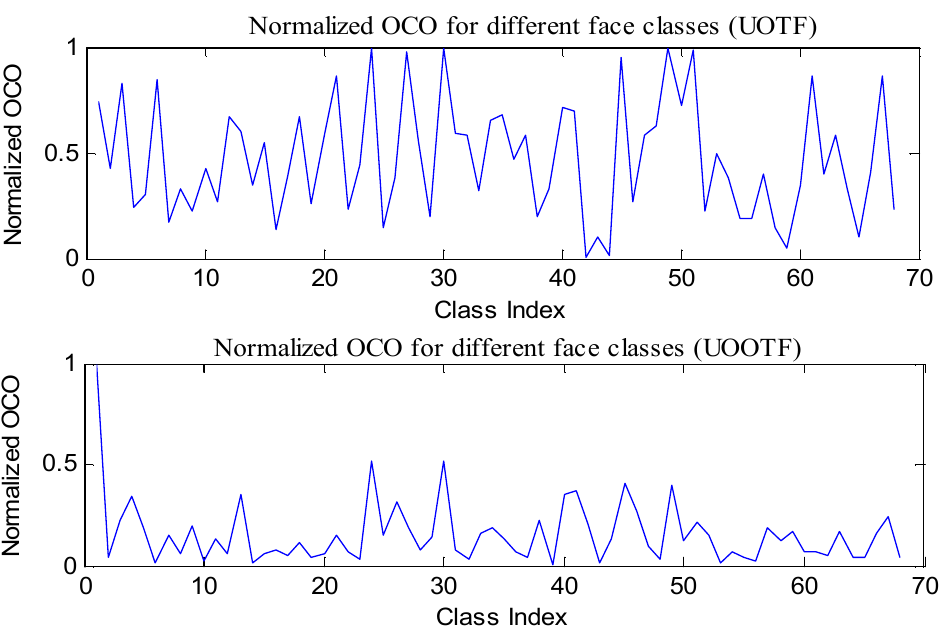}
         \caption{Normalized Origin Correlation Outputs (OCO) for different face classes on the PIE database. Top: Normalized OCO for UOTF; Bottom: Normalized OCO for UOOTF.
                 }
         \label{FIG:Compare}%
        \end{figure*}

It is worth comparing the performance obtained by different types of unconstrained correlation filters, which are designed based on various optimization criteria. The traditional unconstrained correlation filter, such as UOTF, is designed based on the overall correlation output plane. Nevertheless, such kind of filter design is not consistent with the feature extraction process, where only the origin correlation output is used in 1D-CFA. In contrast, during the design of UOOTF, the
optimization criterion only focuses on the origin correlation output which is in essence more appropriate for feature extraction. Fig. \ref{FIG:Compare} shows the normalized origin correlation
outputs (OCO) for UOOTF and UOTF on a test face on the PIE face database \cite{Sim2002}. We can observe in Fig. \ref{FIG:Compare} that UOOTF can produce only one large amplitude peak value (equal to $1$) for the relevant class while suppressing the peak values of the other irrelevant classes. On the contrary, UOTF  produces multiple large amplitude peak values (close to $1$) for several classes due to overfitting.

Different from OTF, UOOTF optimizes the correlation origin outputs for the extra-class samples. In addition, the solution of UOOTF is simpler and the output distortion tolerance is further enhanced by relaxing the constraints on the correlation peaks for the intra-class samples.

However, we should point out that the proposed UOOTF
 requires more training time to obtain the closed-form solution compared with the other filters, such as UOTF and OTF. This is because the non-diagonal matrix inversion, which consumes the majority of the CPU time, is employed in UOOTF (see (\ref{EQ:7})) during the filter design. This problem can be alleviated by considering using GPU or parallel computation \cite{Wagner2011}.
\section{Kernel UOOTF}

In this section, UOOTF is designed in a high-dimensional feature
space by using the kernel technique. The main idea of the kernel correlation filter is to find a non-linear projection by non-linearly mapping the data onto a feature space $\mathcal{F}$ and then design the correlation filter there, thus implicitly yielding a non-linear filter in the input space \cite{Mika1999,Liu2002}.

Suppose $\phi:\vec{\bf{Y}} \in \mathbb{Z}^{p}\to \vec{\bf{F}}\in \mathcal{F}$ is the non-linear mapping. Kernel UOOTF (i.e., KUOOTF) tries to design the filter in $\mathcal{F}$ so as to maximize the following objective function:
\begin{equation}
\label{EQ:9}
J(\vec{\bf{F}}^{l})    =  \frac{\vec{\mathbf{F}}^{l+}\vec{\mathbf{M}}_{\phi}^{l}\vec{\mathbf{M}}_{\phi}^{l+}\vec{\mathbf{F}}^{l}}{{\vec{\bf{F}}^{l+}}(\omega_{s} {\mathbf{R}}_{\mathrm{\phi}}^{l} + \omega_{n}{\mathbf{C}_{\phi}}){\vec{\mathbf{F}}}^{l}},
\end{equation}
where $\vec{\mathbf{M}}_{\phi}^{l}$, ${\mathbf{R}}_{\mathrm{\phi}}^{l}$, and $\mathbf{C}_{\phi}$ are the average correlation height, auto-covariance signal matrix and  auto-covariance noise matrix in the feature space for the $l$-th class, respectively. To be specific,
\begin{eqnarray}
\label{EQ:10}
\vec{\bf{M}}_{\phi}^{l} &=& \frac{1}{N_{l}}\sum_{i=1}^{N_{l}}\phi(\vec{\bf{Y}}_{i}^{\mathrm{I}}),  \nonumber \\
{\mathbf{R}}_{\mathrm{\phi}}^{l} &=& \frac{1}{N_{el}}\sum_{i=1}^{N_{el}}\phi(\vec{\bf{Y}}_{i}^{\mathrm{E}})\phi(\vec{\bf{Y}}_{i}^{\mathrm{E}})^{+}, \nonumber \\
\mathbf{C}_{\phi} &=& \frac{1}{N_{el}}\sum_{i=1}^{N_{el}}\phi(\vec{\bf{N}}_{i}^{\mathrm{E}})\phi(\vec{\bf{N}}_{i}^{\mathrm{E}})^{+}.
\end{eqnarray}

According to the theory of reproducing kernels \cite{Mika1999}, any solution $\vec{\bf{F}}^{l}\in \mathcal{F}$ must lie in the
span of all the training samples in $\mathcal{F}$. Thus, the solution can be expressed as
\begin{equation}
\label{EQ:11}
\vec{\bf{F}}^{l}    = \sum_{i=1}^{N}\alpha^{l}_{i}\phi(\vec{\bf{Y}}_{i}).
\end{equation}

Based on (\ref{EQ:11}) and the expression of $\vec{\bf{M}}_{\phi}^{l}$ in (\ref{EQ:10}), we have
\begin{equation}
\label{EQ:12}
\vec{\mathbf{F}}^{l+}\vec{\mathbf{M}}_{\phi}^{l} = \vec{\alpha}^{l+}\vec{\mathbf{U}}^{l}.
\end{equation}
Here we define $(\vec{\mathbf{U}}_{l})_{i} = 1/N_{l}\sum_{j=1}^{N_{l}}k(\vec{\bf{Y}}_{i},\vec{\bf{Y}}_{j}^{\mathrm{I}})$, where $k(\vec{\bf{X}},\vec{\bf{Y}}) = <\phi(\vec{\bf{X}}),\phi(\vec{\bf{Y}})>$ is the kernel function which only computes
the inner product of two vectors in $\mathcal{F}$ without ever mapping the data explicitly.

Therefore, we have
\begin{equation}
\label{EQ:13}
\vec{\mathbf{F}}^{l+}\vec{\mathbf{M}}_{\phi}^{l}\vec{\mathbf{M}}_{\phi}^{l+}\vec{\mathbf{F}}^{l} = \vec{\alpha}^{l+}\vec{\mathbf{U}}^{l}\vec{\mathbf{U}}^{l+}\vec{\alpha}^{l}.
\end{equation}

Using a similar transformation as (\ref{EQ:13}), we can obtain
\begin{equation}
\label{EQ:14}
{\vec{\bf{F}}^{l+}}(\omega_{s}{\mathbf{R}}_{\mathrm{\phi}}^{l} + \omega_{n}{\mathrm{C}_{\phi}}){\vec{\mathbf{F}}}^{l}
= \vec{\alpha}^{l+}\mathbf{K}^{l}\vec{\alpha}^{l},
\end{equation}
where\begin{eqnarray}
\label{EQ:15}
\mathbf{K}^{l} &=& \omega_{s}(\frac{1}{N_{el}}\sum_{i=1}^{N_{el}}\vec{\bf{\psi}}_{i}^{\mathrm{E}}\vec{\bf{\psi}}_{i}^{\mathrm{E}+})
+ \omega_{n}(\frac{1}{N_{el}}\sum_{i=1}^{N_{el}}\vec{\bf{\upsilon}}_{i}^{\mathrm{E}}\vec{\bf{\upsilon}}_{i}^{\mathrm{E}+}),  \nonumber \\
\vec{\bf{\psi}}_{i}^{\mathrm{E}} &=& (k(\vec{\bf{Y}}_{1},\vec{\bf{Y}}_{i}^{\mathrm{E}}), k(\vec{\bf{Y}}_{2},\vec{\bf{Y}}_{i}^{\mathrm{E}}),\cdots,k(\vec{\bf{Y}}_{N},
\vec{\bf{Y}}_{i}^{\mathrm{E}}))^{+},  \nonumber \\
\vec{\bf{\upsilon}}_{i}^{\mathrm{E}} &=& (k(\vec{\bf{Y}}_{1},\vec{\bf{N}}_{i}^{\mathrm{E}}),k(\vec{\bf{Y}}_{2},\vec{\bf{N}}_{i}^{\mathrm{E}}),\cdots,k(\vec{\bf{Y}}_{N},
\vec{\bf{N}}_{i}^{\mathrm{E}}))^{+}.
\end{eqnarray}

As a result, maximizing (\ref{EQ:9}) is equivalent to maximize
\begin{equation}
\label{EQ:16}
J(\vec{\alpha}^{l}) = \frac{\vec{\alpha}^{l+}\vec{\mathbf{U}}^{l}\vec{\mathbf{U}}^{l+}\vec{\alpha}^{l}}{\vec{\alpha}^{l+}\mathbf{K}^{l}\vec{\alpha}^{l}}.
\end{equation}

The above problem can be solved analogously to (\ref{EQ:6}). Thus, the solution of (\ref{EQ:16}) is
\begin{equation}
\label{EQ:17}
\vec{\alpha}^{l} = (\mathbf{K}^{l})^{-1}\vec{\mathbf{U}}^{l}.
\end{equation}

The training stage of the KUOOTF based 1D-CFA is similar to the UOOTF based 1D-CFA except for the step of computing the feature matrix, which is given in Algorithm 2.

\begin{table}[t]
\centering
\vspace{3mm}
\noindent
\scalebox{0.85}{
\begin{tabular}
{p{348pt}}
\hline
\textbf{Algorithm 2} Computation of the feature matrix for the KUOOTF based 1D-CFA \\
\hline
\vspace{0.05mm}
\textbf{Input:} Low-dimensional PCA feature matrix $\rm{Y} \in$$ \Re^{p\times}$$^N$ with $L$ classes, where $p$ is the
dimensionality of the feature, and the corresponding 1D Fourier transform matrix $\mathbf{Y} \in \Re^{p\times }$$^N$.\\
\textbf{Output:} The feature matrix $\rm{X}$.
\vspace{2mm}
\\
\hline
\vspace{0.1mm}
\emph{Step 1}:
 Do for $l$ = 1,$\cdots$, $L$: \\
 \quad \quad \quad 1.1 Calculate the tradeoff $\mathbf{K}^{l}$ based on the kernel function of the extra-class samples of the $l$-th class;\\
 \quad \quad \quad 1.2 Calculate the average value $\vec{\mathbf{U}^{l}}$  based on the kernel function of the intra-class samples of the $l$-th class;\\
 \quad \quad \quad 1.3 Calculate the weight vector $\vec{\alpha}^{l}$ via (\ref{EQ:17}) of the $l$-th class.\\
\emph{Step 2}: Compute the feature matrix $\rm{X}$ based on (\ref{EQ:18}).\\
\hline
\end{tabular}
}
\end{table}

\vspace{2mm}

During the testing stage, the origin correlation output of a sample $\vec{\bf{Y}}$ using the kernel filter in $\mathcal{F}$ is given by
\begin{equation}
\label{EQ:18}
\vec{\bf{F}}^{l+}\phi(\vec{\bf{Y}})
 = \sum_{i=1}^{N}\alpha^{l}_{i}k(\vec{\bf{Y}}_{i},\vec{\bf{Y}}).
\end{equation}
\section{Experiments}

We evaluate the performance of the proposed UOOTF based 1D-CFA and its kernelization (KUOOTF) in face recognition.
In our experiments, we use the widely-used AR \cite{Martinez1998},
FERET \cite{Phillips1998}, FRGC \cite{Phillips2005}, LFW \cite{Huang2007}, and CAS-PEAL-R1 \cite{Gao2008} face databases.
These face databases contain a wide range of facial variations with different conditions including changes in facial expression (in AR, FERET, LFW, and CAS-PEAL-R1), illumination (in AR, FRGC, LFW, and CAS-PEAL-R1), and pose (in FERET and LFW).

The methods chosen for comparisons are the Eigenface method (PCA) \cite{Turk1991}, the Fisherface method (PCA+LDA) \cite{Belhumeur1997}, the MGMD method \cite{Tao2009}, the Laplacianface method (PCA+LPP) \cite{He2005}, and 1D-CFA (including OTF \cite{Kumar2006} based
and UOTF \cite{Kumar1999} based methods). In addition, the kernel subspace learning methods including the recently proposed eigenspectrum
regularization based kernel LDA (ER-KDA) \cite{Zafeiriou2012} and
the kernel OTF (KOTF) based CFA \cite{Xie2005b}, are also selected.

All the face images are cropped and normalized to the size of $64\times64$. Histogram equalization is applied to the face images for photometric normalization. The linear combination coefficient in MGMD \cite{Tao2009} is chosen by cross-validation in the training set.
The reduced dimensionality of the PCA subspace in 1D-CFA is set to $N-1$ ($N$ is the number of all the training samples).
In our experiments, to demonstrate the capability of feature extraction for different subspace learning methods, the pixel intensities \cite{Turk1991} and Gabor features \cite{Liu2004} are respectively used for representing the face images. In particular, we use the Gabor wavelets with five scales and eight orientations and then down-sample the obtained features by a factor (four in our case).
For the kernel based methods, the widely-used Gaussian RBF kernel
$k(\vec{\bf{X}},\vec{\bf{Y}})= \rm{exp}(-||\vec{\bf{X}}-\vec{\bf{Y}}||^{2}/ \delta^{2})$ is applied. Other kernel functions, such as the polynomial kernel function, could also be used. However, the performance difference by using the two kernels is not significant \cite{Xie2005b,Zafeiriou2012}. Therefore, we mainly focus on the RBF kernel in our experiments.

\subsection{AR Database}
\begin{figure}[tbh!]
         \centering
            \includegraphics[width=7.6cm,height=2.6cm]{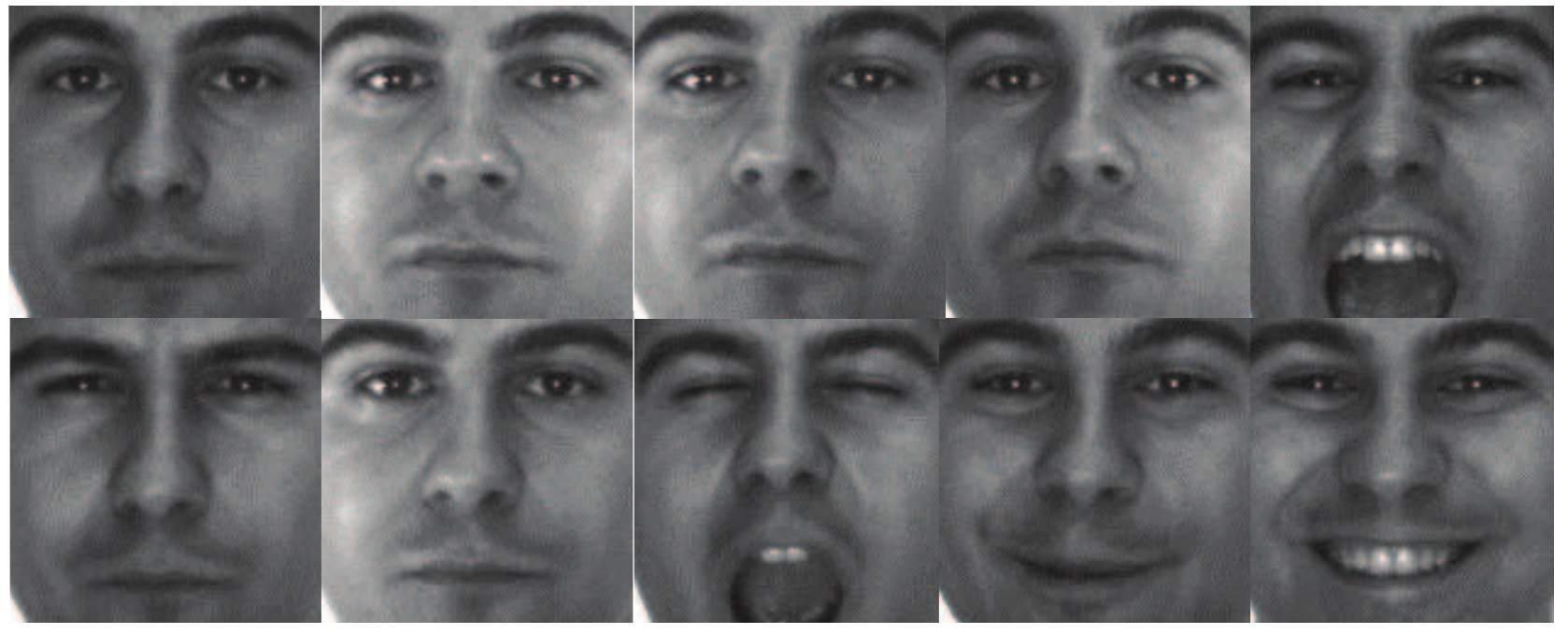}%
         \caption{ Sample images of one person on the AR database.
                 }
         \label{FIG:AR}%
        \end{figure}

The AR database \cite{Martinez1998} consists of more than 4,000 frontal face images of 120 persons. Each person has up to 26 images taken in two sessions. The first session contains 13 images, including different facial expressions, illumination variations, and occlusions. The second session duplicates the first session two weeks later.
We select 14 face images (each session contains seven images) from each of these 120 individuals.
Fig. \ref{FIG:AR} shows the sample images of one person used in our experiments.

A random subset (with $m$ images per individual) is taken from the database to form the training set. The rest of the image database is used for testing. For each $m$, the experiments with randomly sampled subsets are implemented twenty times. We report the top average recognition rate and the corresponding dimensionality of the reduced subspace over the randomly sampled testing sets as the final results \cite{Zhu2012}. Moreover, the highest recognition rate for each case is shown in the bold font. In this paper, we focus on the
small sample size problem, which is one of the most fundamental issues in face recognition \cite{Jain1997,Tan2008,Zhu2012}. Therefore, for all the databases, the value of $m$ is set to 2 and 3.

\begin{figure*}[tbh!]
         \centering
         \subfigure[$m=2$]{
            \includegraphics[width=6.65cm,height=4.6cm]{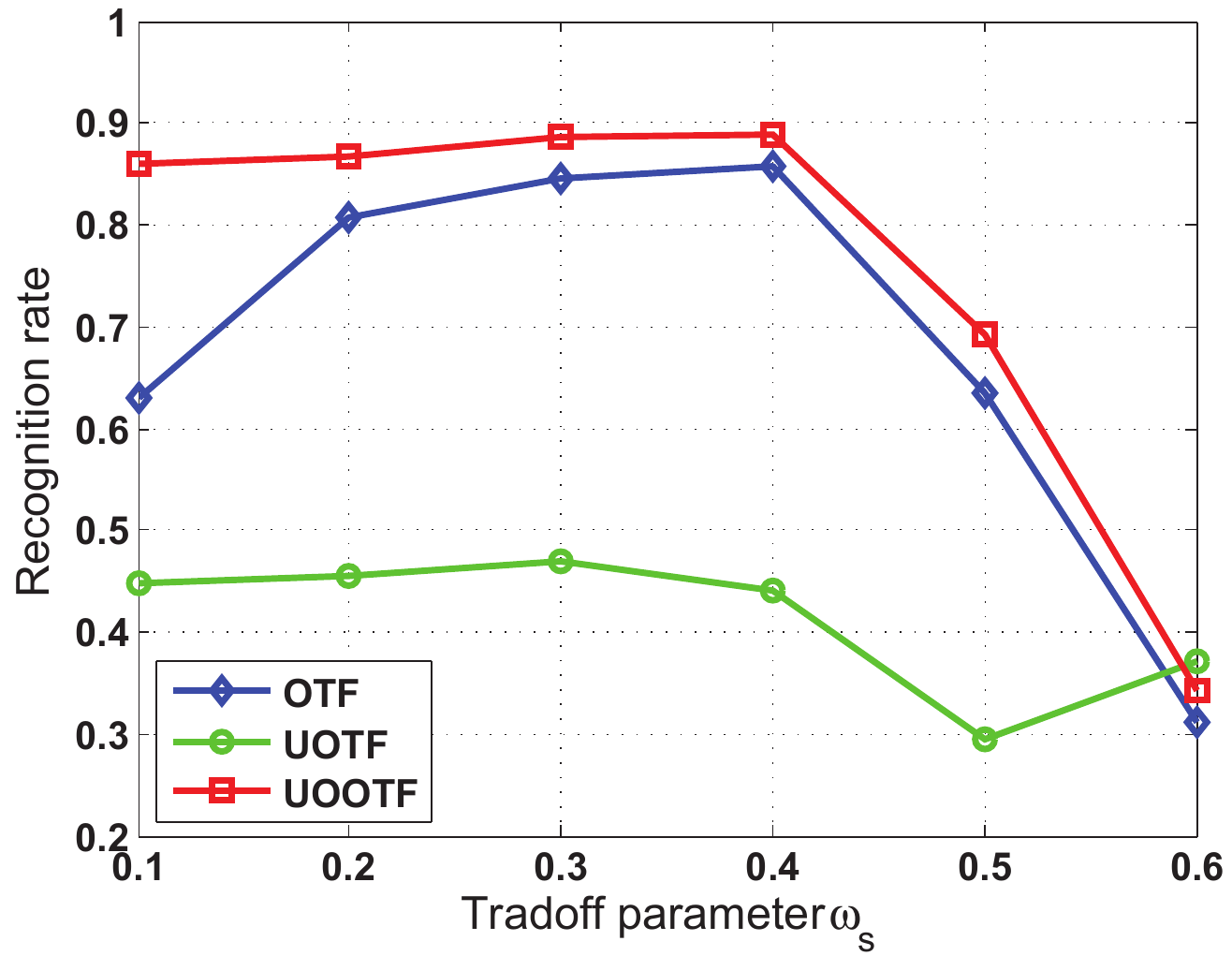}%
            }
          \subfigure[$m=3$]{
            \includegraphics[width=6.65cm,height=4.6cm]{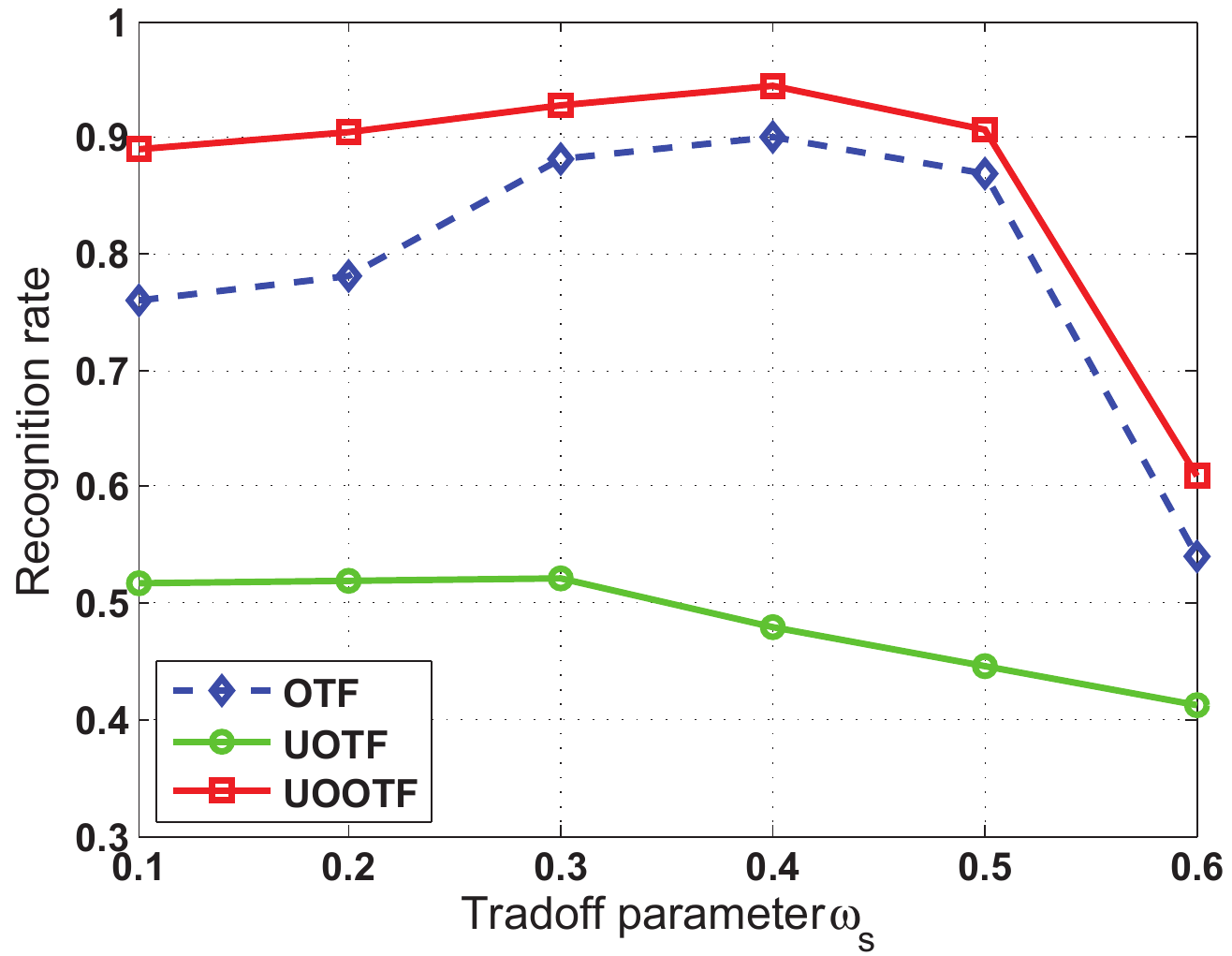}%
            }
         \caption{ Recognition rate vs. tradeoff parameter $\omega_{s}$ under $m=2$ and $m=3$.
                 }
         \label{FIG:Tradeoff}%
        \end{figure*}

 In our experiments, we set the tradeoff parameter $\omega_{n}$ equal to $\sqrt{1-\omega_{s}^{2}}$ (which is same as \cite{Xie2005a,Kumar2006,Xie2005b}).
 We test different settings of the tradeoff parameter $\omega_{s}$.
 In the cases of $m=2$ and $m=3$, the recognition rates based on the pixel intensities vary with different values of $\omega_{s}$, which are respectively shown in Fig. \ref{FIG:Tradeoff} (a) and (b).

The recognition rates achieve the best results when $\omega_{s}$ is 0.4 for OTF and UOOTF.
Meanwhile, the optimal $\omega_{s}$
for UOTF is 0.3. In fact, we observe similar results on other face
databases. Therefore, we set $\omega_{s}$ to be constant values ($\omega_{s} = 0.4$ for OTF and UOOTF/KUOOTF; $\omega_{s} = 0.3$ for UOTF) for all the following experiments.

\begin{figure*}[tbh!]
         \centering
         \subfigure[$m=2$]{
            \includegraphics[width=6.65cm,height=4.6cm]{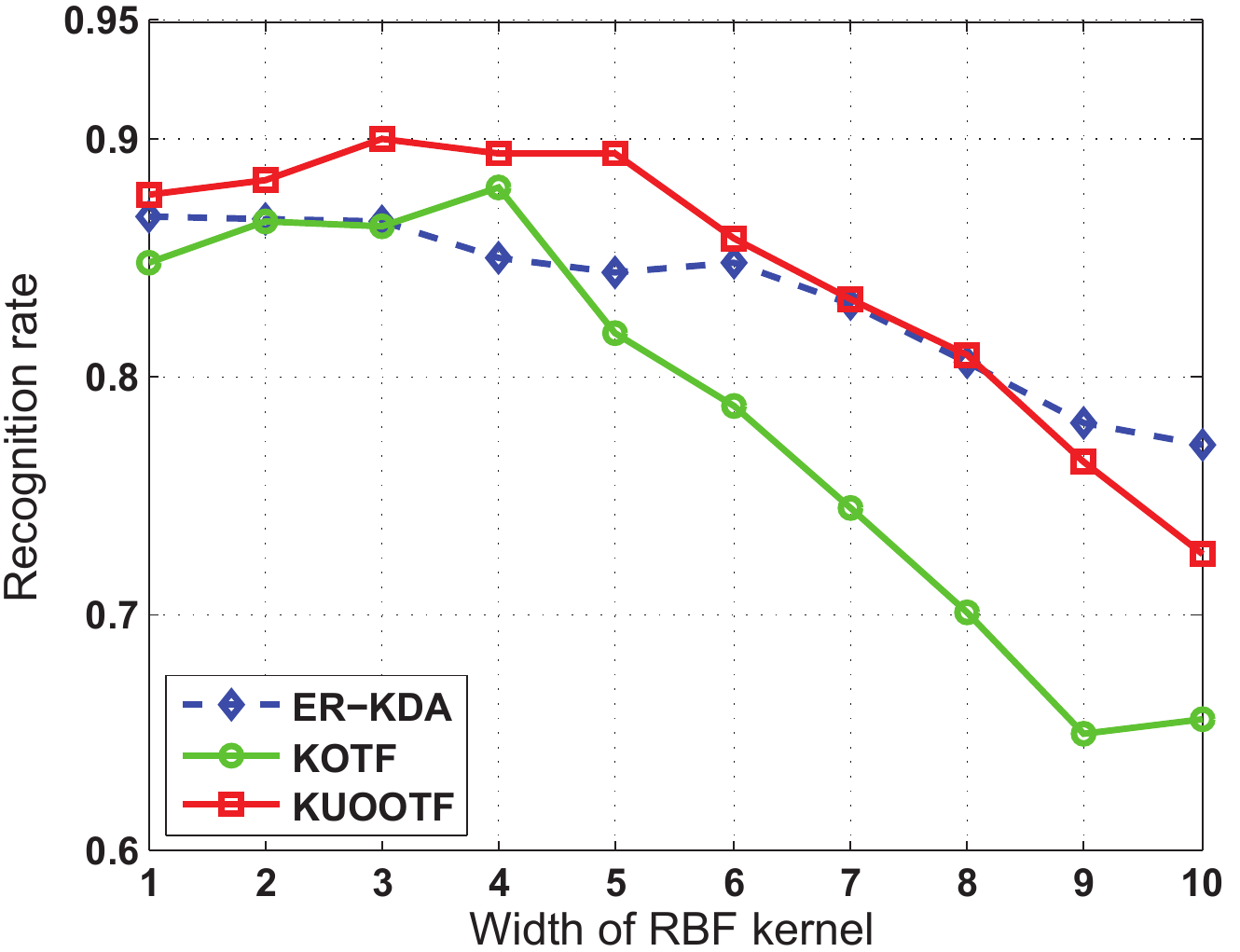}%
            }
          \subfigure[$m=3$]{
            \includegraphics[width=6.65cm,height=4.6cm]{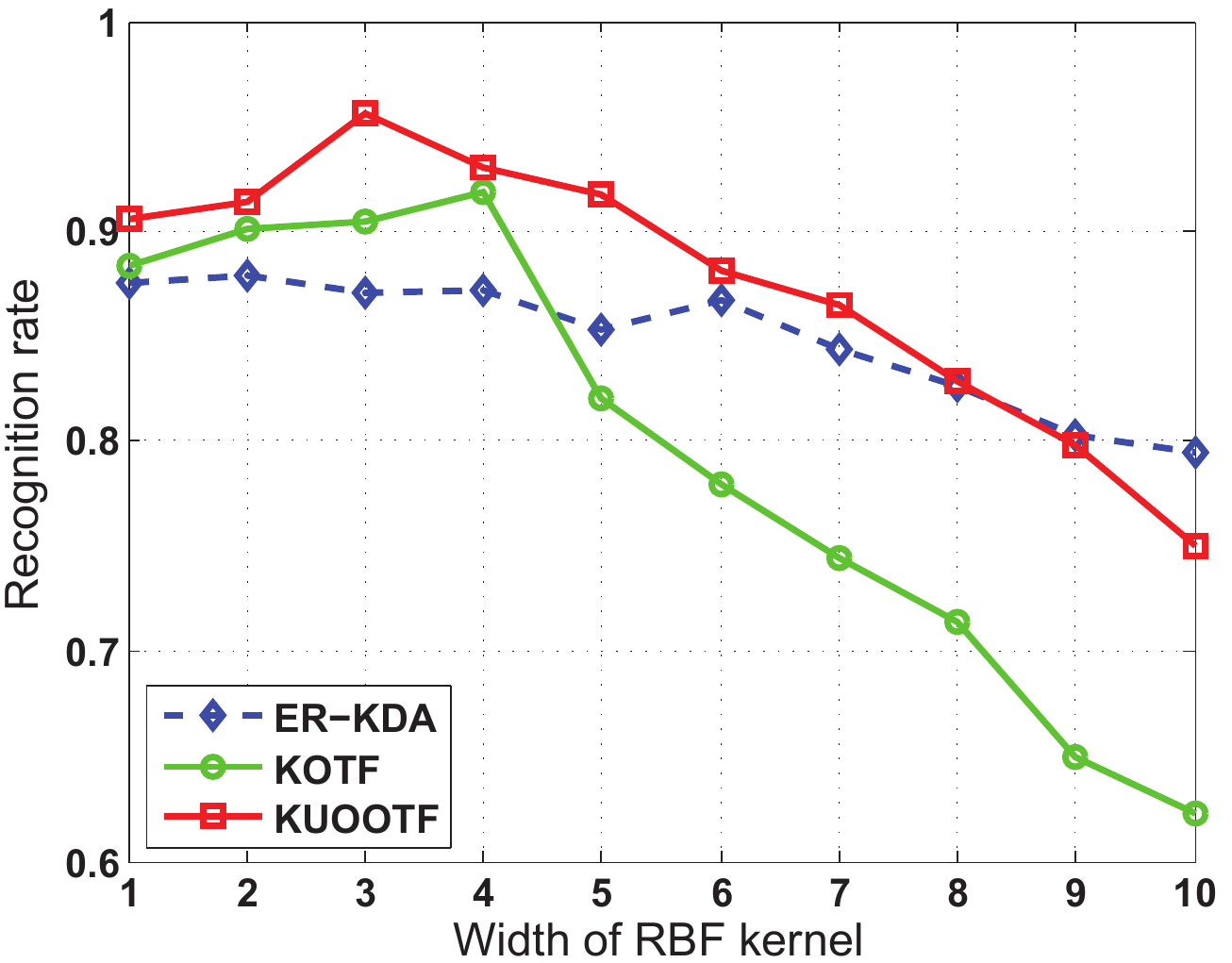}%
            }
         \caption{ Recognition rate vs. the width of RBF kernel under $m=2$ and $m=3$.
                 }
         \label{FIG:KernelCurve}%
        \end{figure*}

To determine proper parameters for kernels, we use the global-to-local search strategy,
which is similar to \cite{Yang2005}. After searching over a wide range of the parameter space, we locate the interval within which the optimal parameters exist. For the Gaussian RBF kernel, the interval is chosen from 1 to 10. The optimal kernel parameters are then found within the interval. Fig. \ref{FIG:KernelCurve}
gives the changes of recognition rates based on the pixel intensities with the different widths of RBF kernel when $m=2$ and $m=3$, respectively.
We can see that the KUOOTF based 1D-CFA achieves better results
than the KOTF based 1D-CFA and ER-LDA.

We experimentally choose the proper kernel parameters which give the results in Fig. \ref{FIG:KernelCurve}. For instance, the width of RBF kernel can be set to 3 for the KUOOTF based 1D-CFA with respect to a
nearest neighbor classifier while the optimal width is 4 for the KOTF based 1D-CFA. By using the kernel technique, we see that the nonlinear kernel extension
is beneficial to improve the performance of UOOTF for feature extraction.

Table \ref{tab:AR} shows the performance of the different subspace learning methods based on the pixel intensities and Gabor features under different values of $m$. From Table \ref{tab:AR}, the recognition accuracy of the UOOTF based 1D-CFA is about $3\%\sim4\%$ better than the OTF based 1D-CFA based on the
pixel intensities. The recognition rates are further improved when the Gabor features are used. However, the performance difference between different methods is smaller for the Gabor features compared to the pixel intensities. This is because the Gabor features can extract high-dimensional features that are more tolerant to variations caused by facial expression and illumination than the pixel intensities. As shown in Table \ref{tab:AR}, the KUOOTF based 1D-CFA achieves the best recognition accuracy among all the competing methods. In addition, KOTF and KUOOTF can still increase the recognition rate about 2\% compared with OTF and UOOTF by using the Gabor features.
         \begin{table*}
         \centering
         \caption
         {
         The top average recognition rate (\%) and the corresponding dimensionality of the reduced subspace (in the bracket) on the AR database.
         }
         \scalebox{0.7}{
         \begin{tabular}{c|cccc}
         \toprule
         Method &Intensity ($m=2$)	&Intensity ($m=3$) 	&Gabor ($m=2$)	&Gabor ($m=3$)\\
         \midrule
          \multirow{2}{*}{Eigenface}  &72.37 &82.67 &78.41 &79.32\\
          &(110) &(101) &(236) &(146) \\
          \hline
         \multirow{2}{*}{Fisherface}  &80.53 &83.15 &81.21	&90.16\\
          &(74) &(119) &(74) &(119) \\
         \hline
         \multirow{2}{*}{MGMD}  &81.15 &82.07 &85.54  &88.18\\
          &(118) &(120) &(115) &(120) \\
         \hline
         \multirow{2}{*}{Laplacianface}  &83.01 &84.54 &89.93  &91.02\\
          &(113) &(110) &(119) &(119) \\
         \hline
          1D-CFA  &85.62 &90.12 &90.40	 &92.14 \\
          (OTF) &(120) &(120) &(120) &(120) \\
         \hline
          1D-CFA  &46.87 &52.09 &60.43 &65.98 \\
          (UOTF) &(120) &(120) &(120) &(120) \\
         \hline
          \bf{1D-CFA}  &88.78 &94.42 &91.83 &95.82 \\
          \bf{(UOOTF)} &(120) &(120) &(120) &(120) \\
          \hline
           \multirow{2}{*}{ER-KDA}   &86.81 &87.29 &89.44 &90.50\\
          &(59) &(128) &(83) &(146) \\
          \hline
          1D-CFA &87.97 &91.84 &91.17 &93.29 \\
          (KOTF) &(120) &(120) &(120) &(120) \\
          \hline
          \bf{1D-CFA} &\bf{90.02} &\bf{95.29} &\bf{92.34} &\bf{96.80} \\
          \bf{(KUOOTF)} &(120) &(120) &(120) &(120) \\

         \bottomrule
         \end{tabular}
         }
         \label{tab:AR}
         \end{table*}
\subsection{FERET Database}

\begin{figure*}[tbh!]
         \centering
            \includegraphics[width=6.2cm,height=2.6cm]{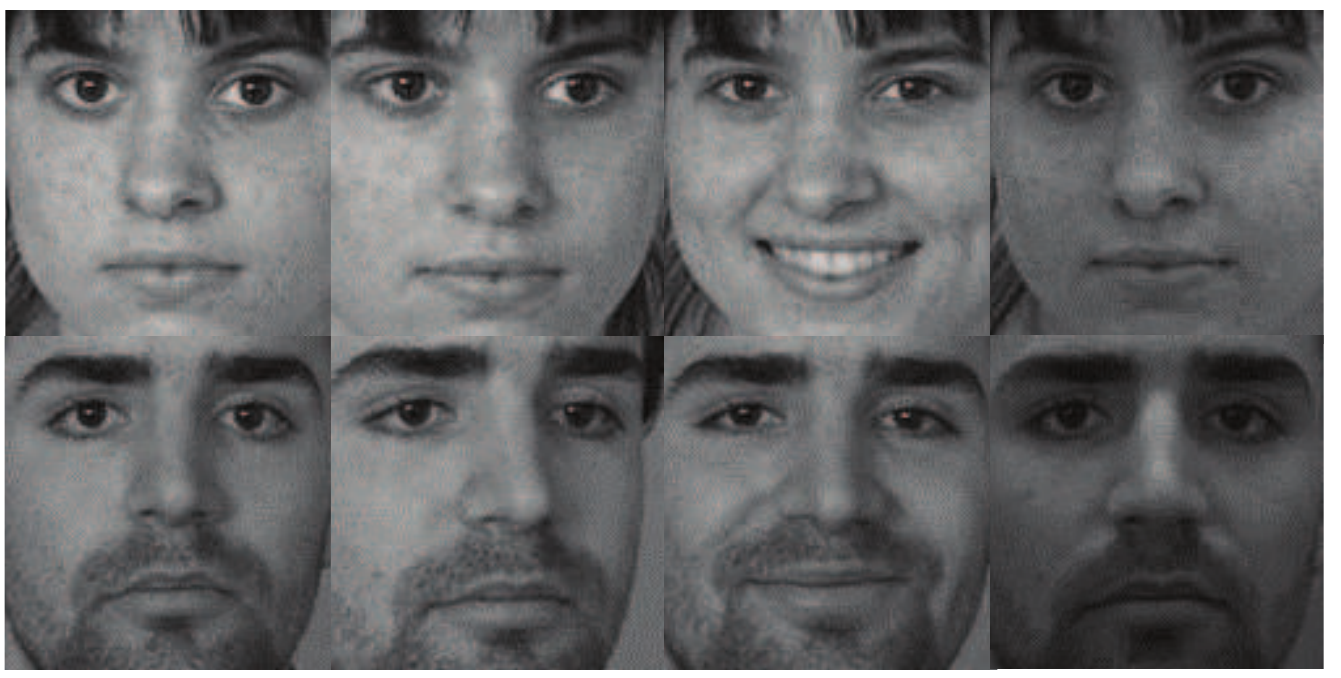}%
         \caption{ Sample images of two persons on the FERET database.
                 }
         \label{FIG:FERET}%
        \end{figure*}

The FERET database \cite{Phillips1998} is a well-known face database for testing and evaluating state-of-the-art face recognition methods. A subset of the FERET database including 800 images and 200 persons (i.e., there are 4 images for each person) is tested. This subset involves variations in facial expression, illumination, and pose. Several examples are given in Fig. \ref{FIG:FERET}.

The recognition results based on the pixel intensities and Gabor features under different values of $m$ are shown
in Table \ref{tab:FERET}.
The performance of UOOTF based 1D-CFA is about 10\% better than the OTF based 1D-CFA with respect to the pixel intensities. The Gabor features help to further improve the recognition accuracy on the FERET. The ER-KDA method performs well on the FERET database, where it achieves the best recognition rate (91.69\%) when $m=2$ and the Gabor features are used. However, the KUOOTF based 1D-CFA obtains the recognition rate up to 99.01\% when more training samples are used (i.e., $m=3$).

\subsection{FRGC Database}

\begin{figure*}[tbh!]
         \centering
            \includegraphics[width=7.3cm,height=2.6cm]{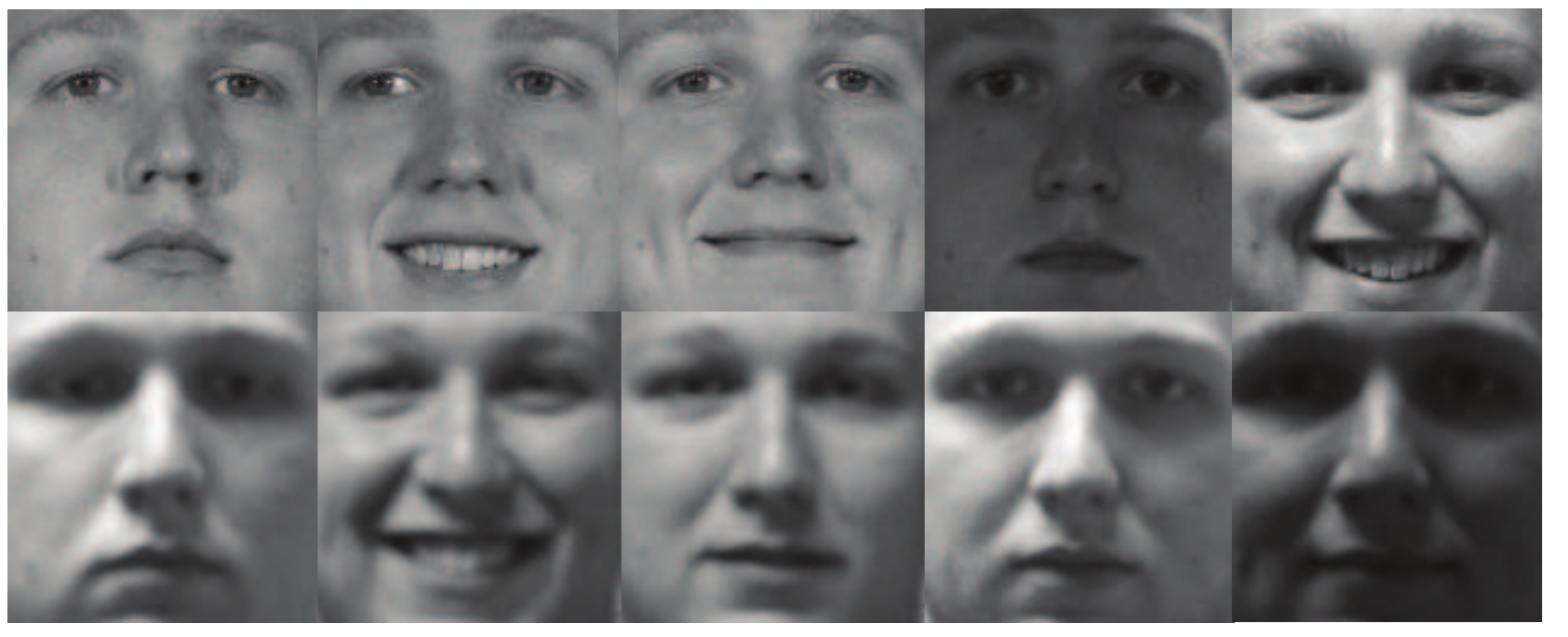}%
         \caption{ Sample images of one person on the FRGC database.
                 }
         \label{FIG:FRGC}%
        \end{figure*}

The FRGC (Face Recognition Grand Challenge) face database \cite{Phillips2005} is another public database for performance evaluation. It consists of controlled images, uncontrolled images and three-dimensional images for each object. We select 6,000 images for 300 individuals (20 images for each person) from the FRGC face database. The face images were captured in both controlled and uncontrolled conditions with harsh illumination and
expression variations. Fig. \ref{FIG:FRGC} shows the sample images of one person used in our experiments.
          \begin{table}
         \centering
         \caption
         {
         The top average recognition rate (\%) and the corresponding dimensionality of the reduced subspace (in the bracket) on the FERET database.
         }
          \scalebox{0.7}{
         \begin{tabular}{c|cccc}
         \toprule
         Method &Intensity ($m=2$)	&Intensity ($m=3$) 	&Gabor ($m=2$)	&Gabor ($m=3$)\\
         \midrule
          \multirow{2}{*}{Eigenface}  &63.90 &66.75 &76.25 &80.22 \\
          &(397) &(595) &(296) &(464) \\
          \hline
         \multirow{2}{*}{Fisherface}  &72.81 &78.63 &83.86 &91.88 \\
          &(82) &(190) &(148) &(199) \\
         \hline
         \multirow{2}{*}{MGMD}  &76.90 &84.22 &86.93  &92.10\\
          &(198) &(200) &(200) &(195) \\
         \hline
         \multirow{2}{*}{Laplacianface}  &74.69 &82.15 &86.41 &93.17 \\
          &(199) &(199) &(199) &(199) \\
         \hline
          1D-CFA  &75.07 &82.25 &89.76 &95.88 \\
          (OTF) &(200) &(200) &(200) &(200) \\
         \hline
          1D-CFA  &27.69 &38.68 &54.00 &59.61 \\
          (UOTF) &(200) &(200) &(200) &(200) \\
         \hline
          \bf{1D-CFA}  &84.95 &93.10 &90.90 &98.25 \\
          \bf{(UOOTF)} &(200) &(200) &(200) &(200) \\
          \hline
          \multirow{2}{*}{ER-KDA} &83.27 &90.17 &\bf{91.69} &95.23 \\
          &(37) &(102) &(129) &(84)\\
           \hline
          1D-CFA &80.25 &85.12 &90.36 &96.12 \\
          (KOTF) &(200) &(200) &(200) &(200) \\
          \hline
          \bf{1D-CFA} &\bf{85.72} &\bf{95.98} &91.24 &\bf{99.01} \\
          \bf{(KUOOTF)} &(200) &(200) &(200) &(200) \\

         \bottomrule
         \end{tabular}
         }
         \label{tab:FERET}
         \end{table}

         \begin{table*}[tbh!]
         \centering
         \caption
         {
         The top average recognition rate (\%) and the corresponding dimensionality of the reduced subspace (in the bracket) on the FRGC database.
         }
          \scalebox{0.7}{
         \begin{tabular}{c|cccc}
         \toprule
         Method &Intensity ($m=2$)	&Intensity ($m=3$) 	&Gabor ($m=2$)	&Gabor ($m=3$)\\
         \midrule
          \multirow{2}{*}{Eigenface}  &47.38 &57.56 &54.68 &63.22 \\
          &(127) &(136) &(392) &(146) \\
          \hline
         \multirow{2}{*}{Fisherface}  &47.89 &53.42 &57.99 &71.43 \\
          &(148) &(199) &(104) &(14) \\
         \hline
         \multirow{2}{*}{MGMD}  &49.45 &57.16 &60.47  &78.14\\
          &(295) &(300) &(300) &(298) \\
         \hline
         \multirow{2}{*}{Laplacianface}  &53.31 &61.21 &68.34 &80.26 \\
          &(299) &(298) &(288) &(294) \\
         \hline
          1D-CFA  &54.32 &62.03 &66.94 &80.91 \\
          (OTF) &(300) &(300) &(300) &(300) \\
         \hline
          1D-CFA  &25.43 &30.61 &40.29 &49.35 \\
          (UOTF) &(300) &(300) &(300) &(300) \\
         \hline
          \bf{1D-CFA}  &64.89 &76.96 &67.27 &87.62 \\
          \bf{(UOOTF)} &(300) &(300) &(300) &(300) \\
          \hline
          \multirow{2}{*}{ER-KDA} &60.07 &62.23 &\bf{68.51} &83.54 \\
          &(82) &(190) &(104) &(127) \\
          \hline
          1D-CFA &55.19 &63.37 &67.23 &84.52\\
          (KOTF) &(300) &(300) &(300) &(300)  \\
          \hline
          \bf{1D-CFA} &\bf{66.43} &\bf{78.39} &67.97 &\bf{88.15} \\
          \bf{(KUOOTF)} &(300) &(300) &(300) &(300) \\
         \bottomrule
         \end{tabular}
         }
         \label{tab:FRGC}
         \end{table*}
The recognition results based on the pixel intensities and Gabor features under different values of $m$ are shown
in Table \ref{tab:FRGC}. From Table \ref{tab:FRGC}, the recognition rates of all the methods are low on the difficult FRGC database. However, UOOTF and KUOOTF can still extract effective features for classification. Besides, we also observe that the Laplacianface achieves slightly better performance than the UOOTF and KUOOTF based 1D-CFA methods on FRGC when we use the Gabor features ($m=2$). But the recognition rates obtained by UOOTF and KUOOTF are significantly higher than the other competing methods when $m=3$.
\subsection{LFW Database}
\begin{figure*}[tbh!]
         \centering
            \includegraphics[width=7.3cm,height=2.6cm]{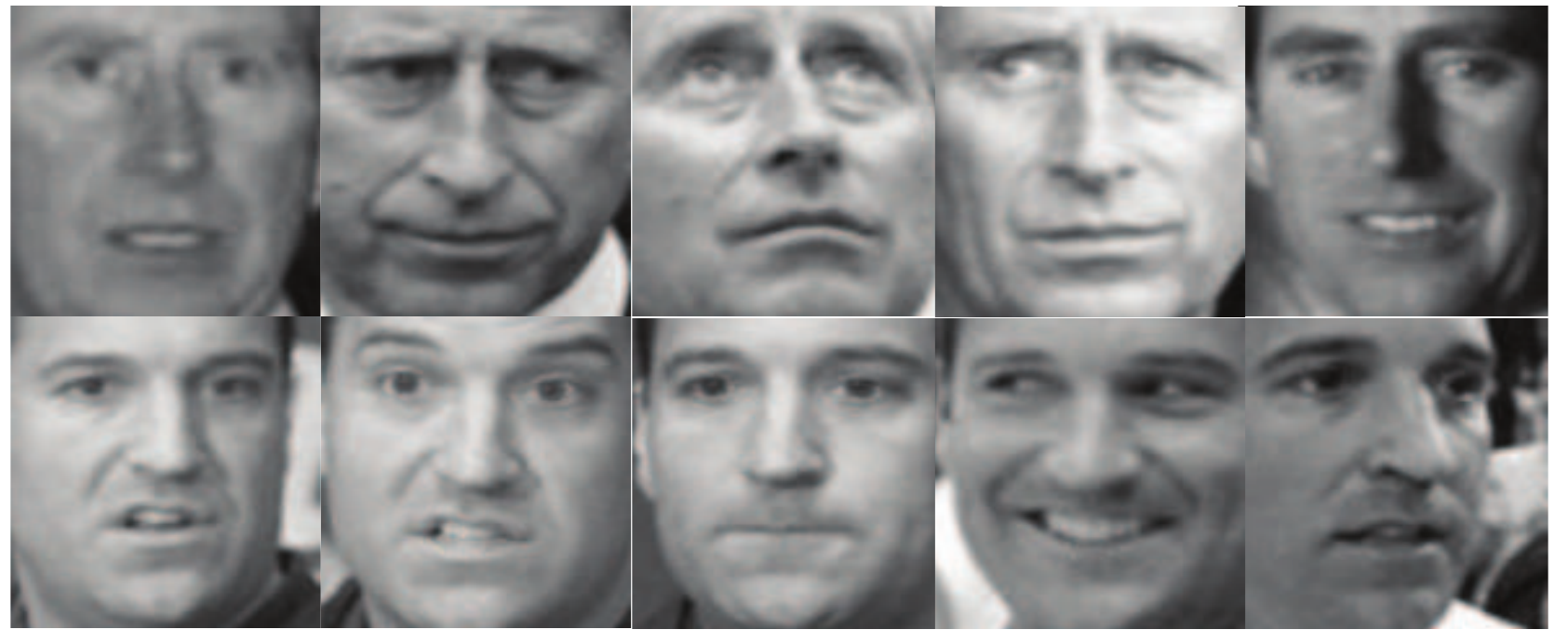}%
         \caption{ Sample images of two persons on the LFW database.
                 }
         \label{FIG:LFW}%
        \end{figure*}

The LFW (Labeled Faces in the Wild) face database \cite{Huang2007} contains images of 5,749 different individuals collected from the web. LFW-a \cite{Wolf2010} is a version of LFW after alignment using a commercial face alignment software. A subset of 100 individuals (5 images for each person) was chosen from the LFW-a database. This subset involves severe variations in illumination, pose, facial expression, age, etc. Fig. \ref{FIG:LFW} shows the sample images of two persons used in our experiments.

The recognition results based on the pixel intensities and Gabor features under different values of $m$ are listed
in Table \ref{tab:LFW}. One can see that the performance of all the methods decreases largely under the unconstrained environments, which demonstrates the difficulty of LFW.
The proposed methods show superior performance to all the other competing methods. Compared with the OTF based 1D-CFA,
the recognition rate is greatly improved by the UOOTF based method. KUOOTF only leads to a little improvement (about 1\%) over UOOTF in LFW.
This is mainly due to the kernel parameter of KUOOTF is fixed for all the databases in our experiments. The kernel parameter of KUOOTF can be further optimized (e.g. using cross validation) for the LFW database.

         \begin{table*}[tbh!]
         \centering
         \caption
         {
         The top average recognition rate (\%) and the corresponding dimensionality of the reduced subspace (in the bracket) on the LFW database.
         }
          \scalebox{0.7}{
         \begin{tabular}{c|cccc}
         \toprule
         Method &Intensity ($m=2$)	&Intensity ($m=3$) 	&Gabor ($m=2$)	&Gabor ($m=3$)\\
         \midrule
          \multirow{2}{*}{Eigenface}  &8.23 &12.28 & 18.09 & 22.46 \\
          &(134) &(245) &(78) &(109) \\
          \hline
         \multirow{2}{*}{Fisherface}  &16.10 & 22.31 & 26.00 & 32.34\\
          &(99) &(97) &(32) &(45) \\
         \hline
         \multirow{2}{*}{MGMD}  &18.34 & 20.57 & 26.20 & 28.45\\
          &(100) &(99) &(100) &(97) \\
         \hline
         \multirow{2}{*}{Laplacianface}  &3.35 & 7.28 & 10.09 & 12.48  \\
          &(90) &(99) &(99) &(99) \\
         \hline
          1D-CFA  &17.14 &  21.19 & 26.00 & 32.34 \\
          (OTF) &(100) &(100) &(100) &(100) \\
         \hline
          1D-CFA  &6.40 & 10.76  &  11.11 & 14.90 \\
          (UOTF) &(100) &(100) &(100) &(100) \\
         \hline
          \bf{1D-CFA}  &25.11 & \bf{32.25} & 35.12 & 42.18\\
          \bf{(UOOTF)} &(100) &(100) &(100) &(100) \\
          \hline
          \multirow{2}{*}{ER-KDA} &23.25 & 26.94 & 32.16 & 37.33\\
          &(56) &(89) &(32) &(12) \\
          \hline
          1D-CFA &20.09 & 28.26  & 27.61 & 35.90\\
          (KOTF) &(100) &(100) &(100) &(100)  \\
          \hline
          \bf{1D-CFA} &\bf{25.85} & 32.19 & \bf{37.00} & \bf{42.78} \\
          \bf{(KUOOTF)} &(100) &(100) &(100) &(100) \\
         \bottomrule
         \end{tabular}
         }
         \label{tab:LFW}
         \end{table*}

\subsection{CAS-PEAL-R1 Database}

\begin{figure*}[tbh!]
         \centering
            \includegraphics[width=6.2cm,height=2.6cm]{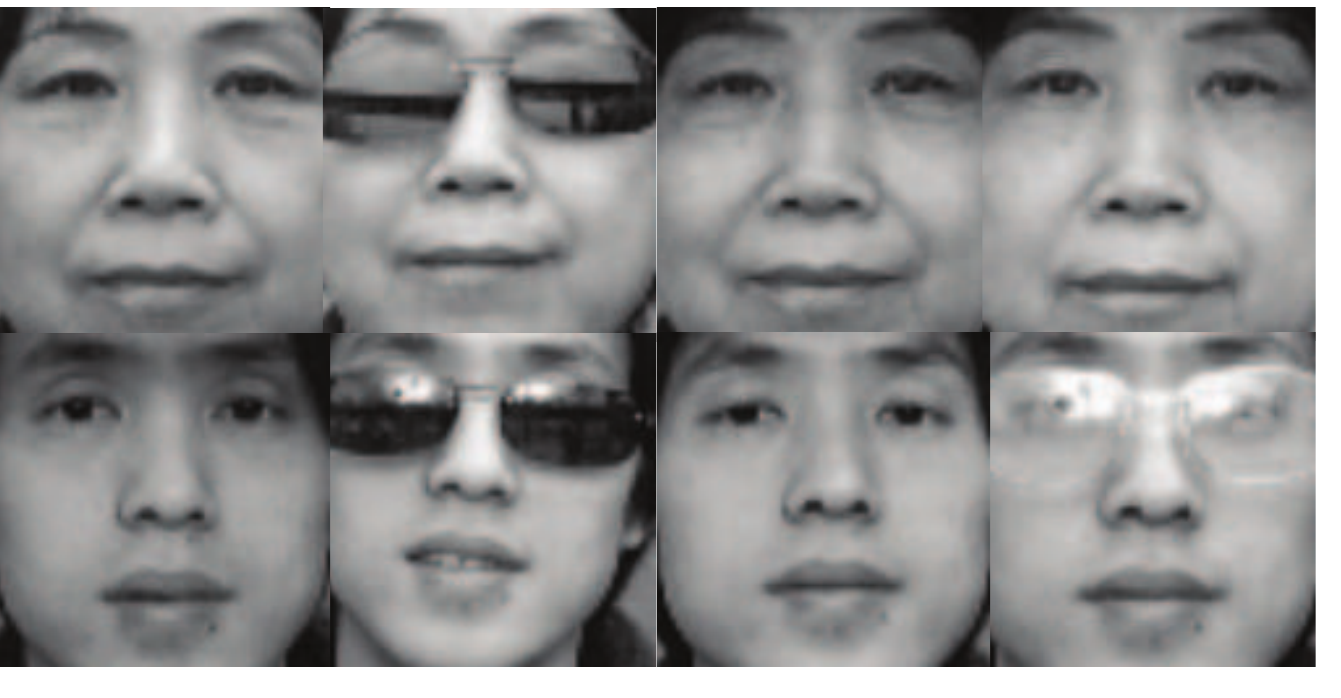}%
         \caption{ Sample images of two persons used for training on the CAS-PEAL-R1 database.
                 }
         \label{FIG:PEAL}%
        \end{figure*}

To test the generalization capability of the proposed methods, we use the CAS-PEAL-R1 Chinese face database and the evaluation protocols introduced in \cite{Gao2008}. The CAS-PEAL-R1 database contains three types of data, i.e., the training set, gallery set and probe set. The training set consists of 1,200 images. The gallery set includes 1,040 images of 1,040 persons (i.e, one image for each person). The CAS-PEAL-R1 database contains six probe sets which correspond to six subsets under different conditions: accessory, age, background, distance, expression, and lighting. All the images that appear in the training set are excluded
from the probe sets and the probe identity may not be trained in the training set. The details of the CAS-PEAL-R1 database are shown in Table \ref{tab:PEALSET}.
Fig. \ref{FIG:PEAL} shows the sample images of two persons used for training.

The recognition rates obtained by the competing methods based on the pixel intensities and Gabor features are shown in Tables \ref{tab:PEAL} and \ref{tab:PEALGabor} respectively, which demonstrate the superiority of our proposed methods on the CAS-PEAL-R1 database under different probe sets. Especially, the KUOOTF based 1D-CFA achieves the approximate 40\% recognition rate under the most difficult lighting set based on the pixel intensities.  The Gabor features further boost the recognition performance for all the subspace learning methods. For the age and the distance probe sets, the recognition rates of the KUOOTF based 1D-CFA are 100\% while the improvement goes up to 56.26\% for the lighting probe set.
       \begin{table*}[tbh!]
         \centering
         \caption
         {
           The datasets used in the CAS-PEAL-R1 database.
         }
          \scalebox{0.67}{
         \begin{tabular}{c|c|c|cccccc}
         \toprule
         \multirow{2}{*}{Datasets} &Training &Gallery &\multicolumn{6}{c}{Probe set}\\
         \cline{4-9}
           &set &set &Accessory	&Age 	&Background & Distance & Expression &Lighting \\
         \midrule
         No. of Images &1,200	&1,040	&2,285	&66	&553 &275 &1,570 &2,243\\
         \bottomrule
         \end{tabular}
         }
         \label{tab:PEALSET}
         \end{table*}
        \begin{table*}[tbh!]
         \centering
         \caption
         {
         The top recognition rate (\%) and the corresponding dimensionality of the reduced subspace (in the bracket) on the CAS-PEAL-R1 database (based on the pixel intensities).
         }
         \scalebox{0.66}{
         \begin{tabular}{c|ccccccc}
         \toprule
         Method &Accessory	&Age 	&Background & Distance & Expression &Lighting & Average\\
         \midrule
          \multirow{2}{*}{Eigenface}  &59.39 &57.58 &95.84 &93.09 &73.69 &10.16 & \multirow{2}{*}{51.00} \\
          &(158) &(56) &(64) &(66) & (139) & (64) \\
         \hline
         \multirow{2}{*}{Fisherface}  &45.95 &33.33 &87.70 &77.45 &61.34 &4.95 & \multirow{2}{*}{40.67} \\
          &(298) &(164) &(209) &(179) & (238) & (144)\\
         \hline
         \multirow{2}{*}{MGMD}  &44.68 & 28.79 &88.43 & 78.91&61.78 &6.78 & \multirow{2}{*}{41.02} \\
          &(300) &(299) &(300) &(290) & (294) & (300)\\
         \hline
         \multirow{2}{*}{Laplacianface}  &38.38 &25.76 &82.28 &70.91 &51.08 &3.30 & \multirow{2}{*}{34.61} \\
          &(319) &(317) &(320) &(294)  &(315) &(265) \\
         \hline
          1D-CFA  &53.52 &56.06 &94.58 &92.00 &67.83 &15.78 & \multirow{2}{*}{49.41} \\
          (OTF) &(300) &(300) &(300) &(300)  &(300) &(300)\\
         \hline
          1D-CFA  &13.74 &6.06 &57.14 &44.00 &41.34 &0.62 & \multirow{2}{*}{13.74} \\
          (UOTF) &(300) &(300) &(300) &(300)  &(300) &(300)\\
         \hline
          \bf{1D-CFA}  &74.84 &71.21 &98.19 &\bf{98.55} &83.12 &31.43 & \multirow{2}{*}{65.52} \\
          \bf{(UOOTF)} &(300) &(300) &(300) &(300)  &(300) &(300)\\
         \hline
          \multirow{2}{*}{ER-KDA} &72.60 &66.67 &97.47 &96.00 &85.29 &20.37& \multirow{2}{*}{61.53} \\
         &(231) &(234) &(325) &(124)  &(132) &(178) \\
         \hline
          1D-CFA &60.96 &63.64 &96.38 &92.73 &75.48 &15.34 & \multirow{2}{*}{53.66} \\
          (KOTF) &(300) &(300) &(300) &(300)  &(300)&(300) \\
          \hline
          \bf{1D-CFA} &\bf{78.29} &\bf{75.76} &\bf{98.37} &\bf{98.55} &\bf{88.09} &\bf{39.05} & \multirow{2}{*}{\bf{70.26}} \\
          \bf{(KUOOTF)} &(300) &(300) &(300) &(300) &(300) &(300) \\
         \bottomrule
         \end{tabular}
         }
         \label{tab:PEAL}
         \end{table*}
        \begin{table*}[tbh!]
         \centering
         \caption
         {
         The top recognition rate (\%) and the corresponding dimensionality of the reduced subspace (in the bracket) on the CAS-PEAL-R1 database (based on the Gabor features).
         }
         \scalebox{0.66}{
         \begin{tabular}{c|ccccccc}
         \toprule
         Method &Accessory	&Age 	&Background & Distance & Expression &Lighting & Average\\
         \midrule
          \multirow{2}{*}{Eigenface}  &67.00 &69.70 &94.03 &94.18 &68.22 &18.64 & \multirow{2}{*}{54.99} \\
          &(118) &(62) &(92) &(101) & (117) & (119) \\
         \hline
         \multirow{2}{*}{Fisherface}  &78.73 &89.39 &88.79 &86.91 &84.20 &17.70 & \multirow{2}{*}{61.60} \\
          &(299) &(299) &(237) &(299) & (224) & (268)\\
         \hline
         \multirow{2}{*}{MGMD}  &80.31 &92.42 & 89.51 &90.91 &85.67 &20.06 & \multirow{2}{*}{63.44} \\
          &(300) &(287) &(300) &(300) & (296) & (300)\\
         \hline
         \multirow{2}{*}{Laplacianface}  &75.67 &87.88 &84.27 &85.82 &51.27 &22.69 & \multirow{2}{*}{54.39} \\
          &(299) &(241) &(254) &(287)  &(299) &(299) \\
         \hline
          1D-CFA  &84.16 &100 &97.11 &98.91 &90.25 &30.58 & \multirow{2}{*}{70.09} \\
          (OTF) &(300) &(300) &(300) &(300)  &(300) &(300)\\
         \hline
          1D-CFA  &37.86 &25.76 &76.49 &77.45 &60.51 &10.48 & \multirow{2}{*}{38.66} \\
          (UOTF) &(300) &(300) &(300) &(300)  &(300) &(300)\\
         \hline
          \bf{1D-CFA}  &87.22 &100 &99.10 &100 &93.57 &50.38 & \multirow{2}{*}{78.39} \\
          \bf{(UOOTF)} &(300) &(300) &(300) &(300)  &(300) &(300)\\
         \hline
          \multirow{2}{*}{ER-KDA} &78.50 &86.36 &98.19 &97.82 &86.88 &30.58& \multirow{2}{*}{67.40} \\
         &(265) &(124) &(176) &(213)  &(120) &(209) \\
         \hline
          1D-CFA &87.61 &100 &99.10&99.27 &93.69 &42.40 & \multirow{2}{*}{75.96} \\
          (KOTF) &(300) &(300) &(300) &(300)  &(300)&(300) \\
          \hline
          \bf{1D-CFA} &\bf{88.36} &\bf{100} &\bf{99.28} &\bf{100} &\bf{95.03} &\bf{56.26} & \multirow{2}{*}{\bf{80.99}} \\
          \bf{(KUOOTF)} &(300) &(300) &(300) &(300) &(300) &(300) \\
         \bottomrule
         \end{tabular}
         }
         \label{tab:PEALGabor}
         \end{table*}

\subsection{Summary and Discussions}

From Tables \ref{tab:AR}-\ref{tab:PEALGabor}, we can see that the UOOTF and KUOOTF based 1D-CFA achieve the comparable or better recognition accuracy compared with the other competing methods. UOOTF is very effective for extracting the features in the 1D-CFA framework. Furthermore,
the kernel extension (KUOOTF) can effectively improve the recognition performance. Compared with ER-KDA and the KOTF based 1D-CFA, the KUOOTF based 1D-CFA can achieve higher recognition rate by $\sim2\%$ to $5\%$ on average. The kernel extension of UOOTF allows for higher flexibility of the decision boundary due to a wider range of non-linearity properties.

 It is worth noting that the overall recognition ability of the UOOTF based 1D-CFA is much better than the UOTF based 1D-CFA in our experiments. The traditional UOTF based 1D-CFA cannot achieve satisfying recognition rates in face recognition because the design criterion in UOTF is to optimize the whole correlation output plane which will lead to the overfitting problem in 1D-CFA (i.e., producing multiple large amplitude peak values). As a matter of fact, the feature extraction in 1D-CFA only considers the origin correlation outputs. Therefore, the UOOTF based 1D-CFA enhances the discriminative ability of the features by allowing the flexible distortion tolerance.
\section{Conclusions}

In this paper, we present an effective unconstrained correlation filter and apply it to the task of face recognition. By removing the hard constraints during the filter design and emphasizing the origin correlation outputs, UOOTF can extract discriminative face features very effectively for classification. Furthermore, we derive the kernel extension of UOOTF to handle non-linearly separable distributions between different classes. Experimental results on several public face databases show that the proposed UOOTF and KUOOTF methods achieve promising results in face recognition.

\section*{Acknowledgments}
The authors would like to thank the associate editor and the anonymous reviewers for their constructive comments on this paper. This work was supported by the National Natural Science Foundation of China under Grants 61201359, 61170179 and 61202279, the Natural Science Foundation of Fujian Province of China under Grant 2012J05126, the National Defense Basic Scientific Research Program of China, and the Specialized Research Fund for the Doctoral Program of Higher Education of China under Grants 20110121110020 and 20110121110033.

\newpage
\parpic{\includegraphics[width=1in,height=1.25in,clip]{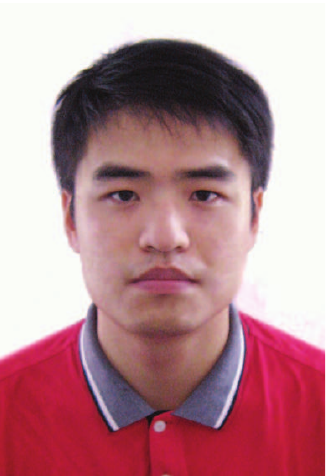}}
\noindent {\bf Yan Yan} is currently an assistant professor in the School of Information Science and Technology at Xiamen University, China. He received the B.S. degree in Electrical Engineering from the University of Electronic Science and Technology of China (UESTC), China, in 2004 and the Ph.D. degree in Information and Communication Engineering from
Tsinghua University, China, in 2009, respectively. He worked at Nokia Japan R\&D center as a research engineer (2009-2010) and Panasonic Singapore Lab as a project leader (2011). His research interests include image recognition and machine learning.
\\
\\
\parpic{\includegraphics[width=1in,height=1.25in,clip]{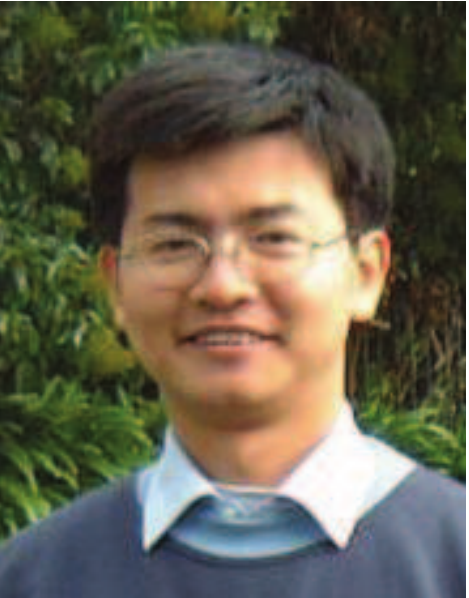}}
\noindent {\bf Hanzi Wang} is currently a Distinguished Professor and "Minjiang Scholar" at Xiamen University, China. He was a Senior
Research Fellow (2008 - 2010) at the
University of Adelaide, Australia; an
Assistant Research Scientist (2007 -
2008) and a Postdoctoral Fellow (2006 -
2007) at the Johns Hopkins University;
and a Research Fellow at Monash University, Australia (2004 -
2006). He received the Ph.D degree in Computer Engineering from
Monash University, Australia. He was awarded the Douglas
Lampard Electrical Engineering Research Prize and Medal for
the best PhD thesis in the Department. His research interests are
concentrated on computer vision and pattern recognition including visual tracking, robust statistics, model fitting, object detection, video segmentation, and related fields. He has published
around 60 papers in major international journals and conferences including the IEEE Transactions on Pattern Analysis and
Machine Intelligence, International Journal of Computer Vision,
ICCV, CVPR, ECCV, NIPS, MICCAI, etc. He is an Associate
Editor for IEEE Transactions on Circuits and Systems for Video
Technology (T-CSVT) and he was a Guest Editor of Pattern
Recognition Letters (September 2009). He is a Senior Member
of the IEEE. He has served as a reviewer for more than 30 journals and conferences.
\\
\\
\parpic{\includegraphics[width=1in,height=1.25in,clip]{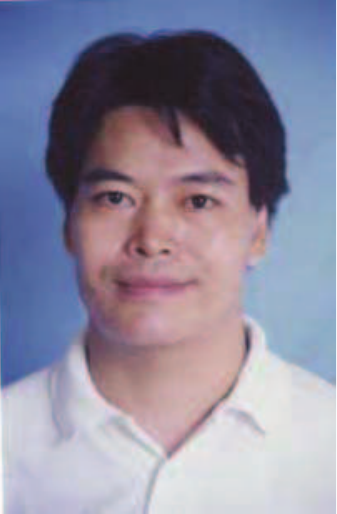}}
\noindent {\bf Cuihua Li} received his Ph.D. degree from Xi'an Jiaotong University, China. He is a professor in the School of Information Science and Technology at Xiamen University, China. His research interests include image processing and image recognition.
\\
\\
\parpic{\includegraphics[width=1in,height=1.25in,clip]{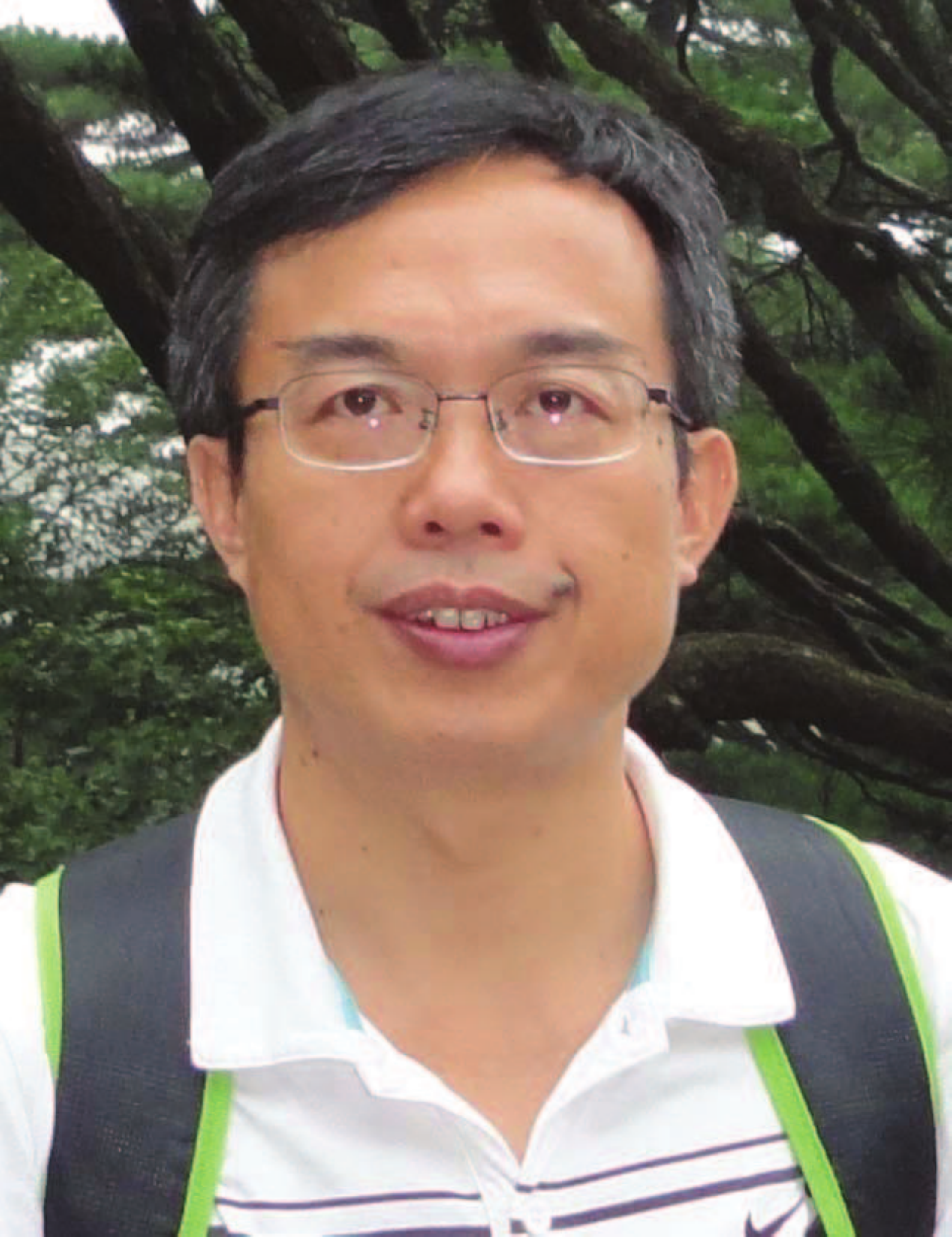}}
\noindent {\bf Chenhui Yang} received his Ph.D. degree from Zhejiang University, China. He is a professor in the School of Information Science and Technology at Xiamen University, China.
His research interests include compute vision and computer graphics.
\\
\\
\parpic{\includegraphics[width=1in,height=1.25in,clip]{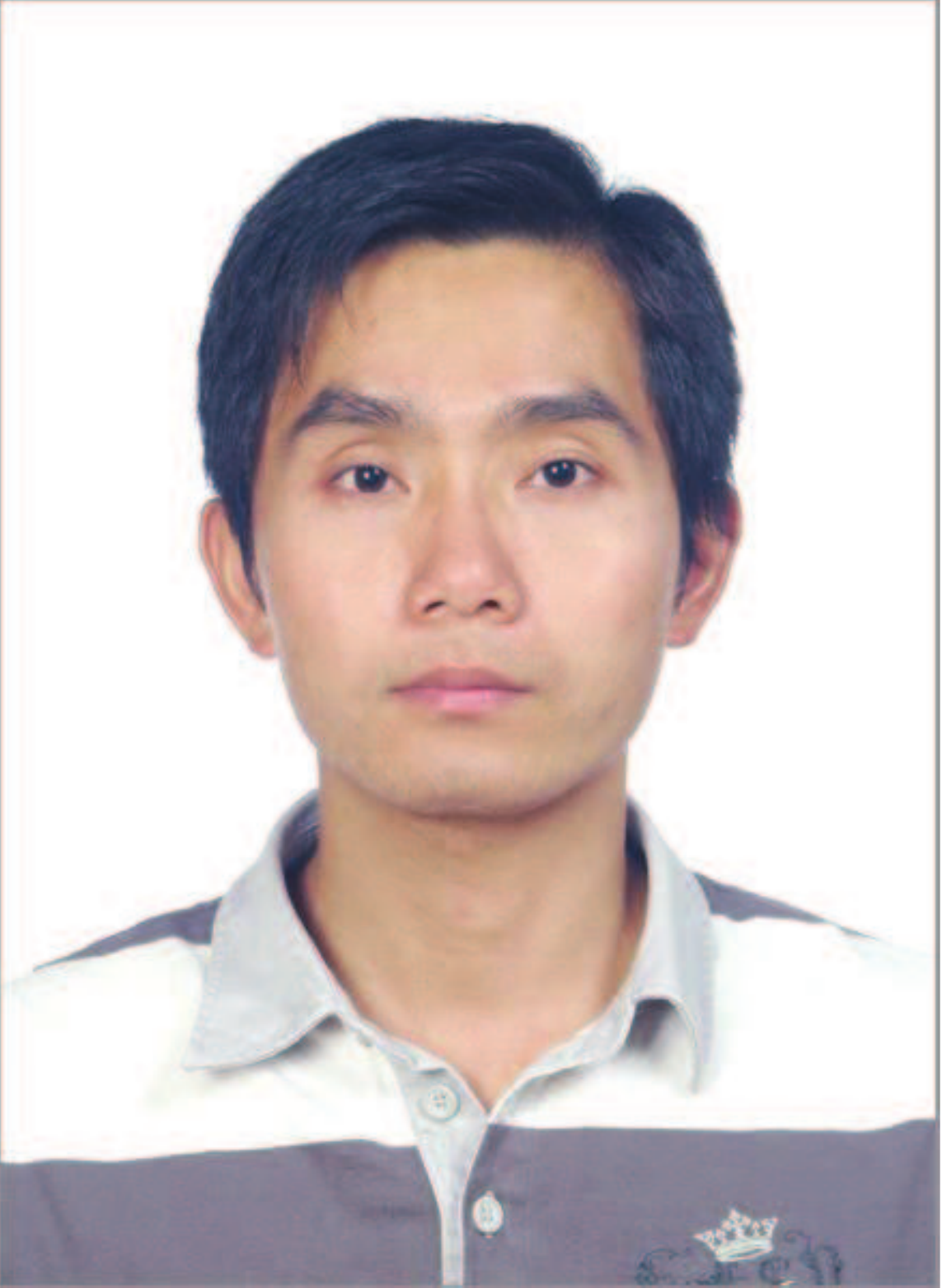}}
\noindent {\bf Bineng Zhong} received the B.S., M.S., and Ph.D. degrees in computer science from Harbin Institute of Technology, China, in 2004, 2006, and 2010, respectively. From 2007 to 2008, he was a Research Fellow with the Institute of Automation and Institute of Computing Technology, Chinese Academy of Science. Currently, he is a Lecturer with the School of Computer Science and Technology, Huaqiao University, China, and he is also a Post-Doc with the School of Information Science and Technology, Xiamen University, China. His current research interests include pattern recognition, machine learning, and computer vision.







\end{document}